\pdfoutput=1
\documentclass[11pt]{article}

\usepackage{acl}
\usepackage{amsmath}
\usepackage{comment}
\usepackage{times}
\usepackage{latexsym}
\usepackage{makecell}
\usepackage{booktabs}    % For professional-looking tables
\usepackage{multirow}    % For multi-row cells
\usepackage{caption}     % For customizing captions
\usepackage{listings}    % For code formatting
\usepackage[table]{xcolor}      % For color customization
\usepackage{tcolorbox}   % For colored boxes (optional, for annotations)
\usepackage{graphicx} 
\usepackage{subcaption} 
\usepackage{paralist}
\usepackage[T1]{fontenc}
\usepackage{float}
\usepackage{placeins}
\usepackage{stmaryrd}
\usepackage{amssymb}
\usepackage{cuted}
\usepackage{bbm}

\usepackage[utf8]{inputenc}

\usepackage{microtype}
\usepackage{inconsolata}

\usepackage{widetext}
\usepackage{flushend}

\newcommand{\best}[1]{\cellcolor{black!20}\textbf{#1}}
\newcommand{\second}[1]{\cellcolor{black!10}\underline{#1}}

\title{Pragmatic Reasoning Improves LLM Code Generation}

\usepackage{authblk}
\author{
  \textbf{Zhuchen Cao}\textsuperscript{1},
  \textbf{Sven Apel}\textsuperscript{2},
  \textbf{Adish Singla}\textsuperscript{3},
  \textbf{Vera Demberg}\textsuperscript{1,2}\\
  \textsuperscript{1}Max Planck Institute for Informatics, Saarland Campus\\
  \textsuperscript{2}Computer Science, Saarland University\\
  \textsuperscript{3}Max Planck Institute for Software Systems, Saarland Campus\\
  \small{\textbf{Correspondence:} \texttt{zcao@mpi-inf.mpg.de, apel@cs.uni-saarland.de, adishs@mpi-sws.org, vera@coli.uni-saarland.de}}
}

\begin{document}
\maketitle
\begin{abstract}

Pragmatic reasoning helps interlocutors infer intended meaning from ambiguous or underspecified messages by considering shared context and counterfactual alternatives. Similar challenges arise in natural language-to-code generation, where user instructions often admit multiple plausible candidate programs. However, direct RSA-style inference is difficult because it requires probability estimation over large spaces of programs and alternative instructions. We propose CodeRSA, an RSA-motivated reranking method that makes pragmatic reasoning tractable through local pragmatic contests among sampled code candidates. CodeRSA constructs candidate-induced alternative instructions and estimates which candidates are most distinctively supported by the original instruction, avoiding global normalization over the full program-instruction space. We evaluate CodeRSA on HumanEval+, MBPP+, and BigCodeBench using four open-weight instruction-following models. CodeRSA achieves the strongest average accuracy in 10 of 12 model-benchmark settings and remains competitive in the remaining cases. Further analyses show that its gains come from combining local pairwise pragmatic comparison with broader global support, suggesting a scalable direction for language-to-code reranking under natural-language uncertainty.

\end{abstract}

\section{Introduction}

Large language models (LLMs) have demonstrated strong performance in generating computer code from natural-language instructions \citep{liu2023your,coignion2024performance}. However, coding tasks can be complex and user instructions are often underspecified, so a single generation attempt may miss correct or higher-quality solutions \citep{liu2024exploring}. A common strategy is to sample multiple solutions, or \emph{code candidates} \citep{chen2021evaluatinglargelanguagemodels,brown2024largelanguagemonkeysscaling}, and then select one through reranking, a strategy central to recent discussions of test-time compute for code generation \citep{li-etal-2025-formalizing}.

\begin{figure*}[t]
\centering
\includegraphics[width=0.9\textwidth]{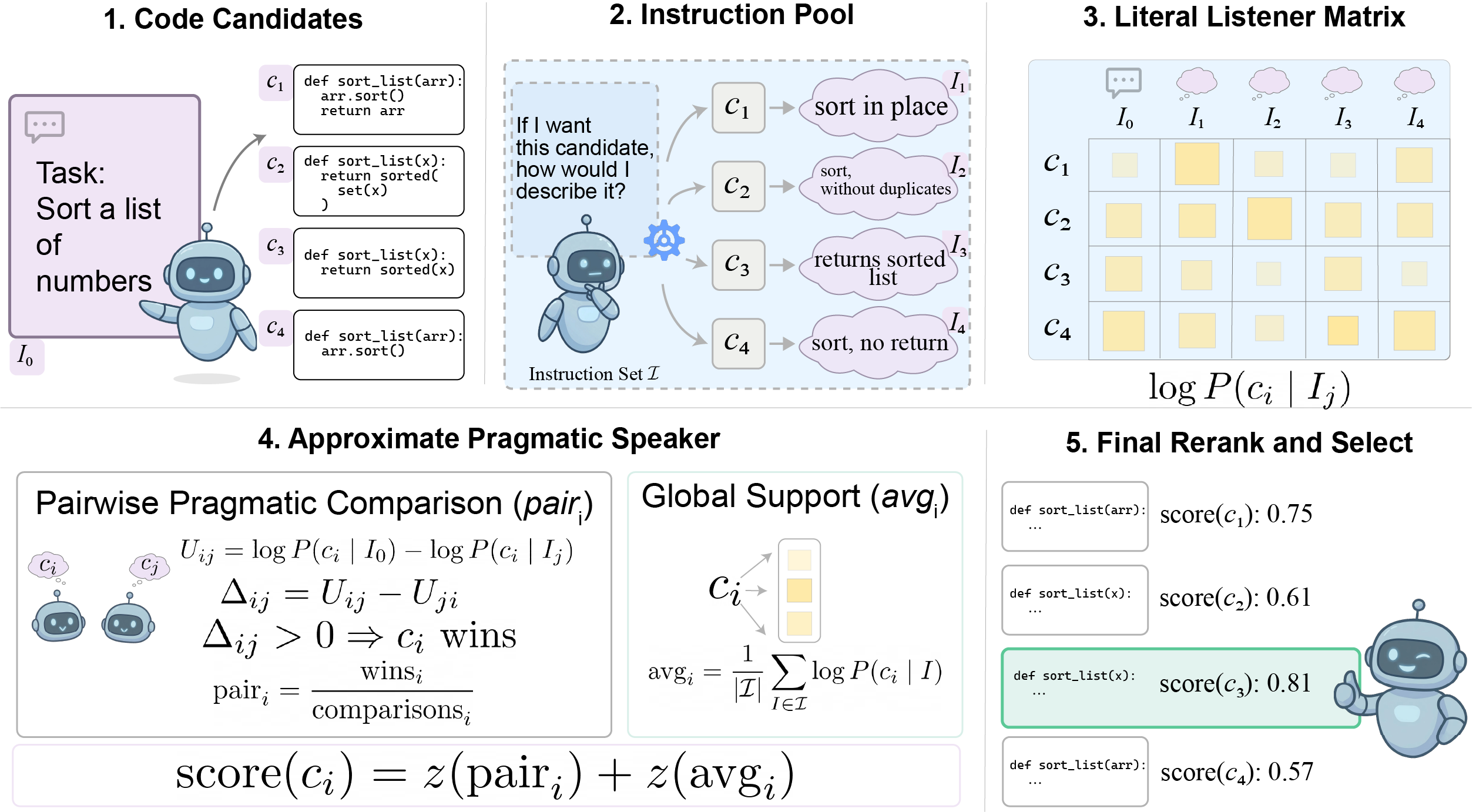}
\caption{Overview of CodeRSA. Given an original instruction $I_0$, we first sample code candidates $c_1,\ldots,c_n$. For each candidate $c_i$, we generate a candidate-induced instruction $I_i$, i.e., a reverse-generated behavior description of what $c_i$ implements. The original instruction and these candidate-induced instructions form a local comparison space. CodeRSA then scores all instruction--candidate pairs with a literal listener, derives local pairwise pragmatic contests and a global-support signal, and combines them for final reranking.}
\label{fig:fig1}
\end{figure*}

Code reranking can be viewed as a pragmatic interpretation problem. In human communication, listeners infer intended meanings from underspecified utterances by considering context and plausible alternatives \citep{grice1975logic}; the Rational Speech Act (RSA) framework formalizes this as recursive reasoning between speakers and listeners \citep{frank2012predicting,goodman2016pragmatic}. In language-to-code generation, the instruction can similarly be treated as an underspecified utterance, and sampled code candidates as possible intended meanings. Candidate selection therefore requires asking not only whether a candidate is plausible under the instruction, but also how well the instruction distinguishes it from relevant alternatives.

Prior work has shown both the promise and difficulty of applying RSA-style reasoning to program generation. Pragmatic reasoning has been effective in regular-expression synthesis \citep{pu2020program,pmlr-v235-pu24c}, but has produced negative results in a spreadsheet setting \citep{schuster2024spreadnala}. A key obstacle is tractability: vanilla RSA would require reasoning over many alternative instructions and code candidates, while the instruction space is effectively open-ended \citep{pmlr-v235-pu24c}. Existing scalable reranking methods therefore often rely on proxy objectives, including direct prompt-conditioned scoring, bidirectional scoring, and inter-candidate consensus \citep{zhang2022coderreviewerrerankingcode,jain-etal-2024-lightweight,jiang2026semanticvotingselfevaluationfreeapproach}. These methods provide useful signals, but do not construct a candidate-induced contrast class for pragmatic comparison.

We propose CodeRSA, an RSA-inspired reranking method for naturalistic language-to-code generation. A straightforward finite approximation to RSA would restrict inference to the sampled candidates and their candidate-induced instructions. However, this induced space is shaped by the sampling process itself: candidates are already plausible under the original instruction, and induced instructions often contain overlapping or paraphrastic descriptions. A single global normalization over this space can therefore amplify redundancy and rely on compressed likelihood differences. CodeRSA instead uses local pairwise pragmatic contests, preserving RSA's contrastive intuition while avoiding dependence on one global normalization over the full induced space.

Starting from a sampled candidate pool, CodeRSA generates one candidate-induced instruction per candidate, forming a task-relevant neighborhood around the original instruction. It then performs local pairwise contests, asking which candidate is more distinctively supported by the original instruction relative to the rival-induced alternative. The resulting tournament-style score is combined with a global-support term that measures plausibility across the induced instruction neighborhood. Together, these signals favor candidates that are both locally distinctive and broadly supported. Figure~\ref{fig:fig1} provides an overview.

We evaluate CodeRSA on HumanEval+ and MBPP+ \citep{chen2021evaluatinglargelanguagemodels,austin2021program,liu2023your}, and BigCodeBench \citep{zhuo2024bigcodebench}, using four open-weight instruction-following models: Llama-3-8B and Llama-3-70B \citep{grattafiori2024llama}, Qwen2.5-7B \citep{qwen25}, and Qwen2.5-32B-Coder \citep{hui2024qwen2}. Across these settings, CodeRSA performs most strongly overall, achieving the best average result in 10 of the 12 combinations of model and benchmark. It consistently improves over Coder and CoderReviewer, and surpasses Consensus-WUCS in 10 of 12 settings \citep{jain-etal-2024-lightweight}. Further analyses show that the pairwise tournament and global-support signals are complementary.

Our main contributions are as follows:
\begin{itemize}
    \item To our knowledge, we are the first to study RSA-inspired local pragmatic comparison for reranking LLM-generated code candidates in naturalistic language-to-code generation.
    \item We propose CodeRSA, a tractable reranker that constructs candidate-induced instruction neighborhoods and selects programs through pairwise pragmatic contests.
    \item We show across three benchmarks and four open-weight models that CodeRSA achieves the best average result in 10 of 12 settings, with complementary gains from local contrast and global support.
\end{itemize}

\section{Related Work}
\label{sec:background}

This section reviews prior work most relevant to CodeRSA and situates our approach within the reranking design space for code generation.

\paragraph{Natural Language to Code.}
Generating code from natural language has been studied extensively, from earlier neural approaches \citep{ling2016latent,rabinovich2017abstract,hayati2018retrieval} to modern large language models based on large-scale pretraining and the transformer architecture \citep{vaswani2017attention}.
Recent LLMs achieve strong performance on code-generation benchmarks and, in some settings, approach or exceed the level of average human programmers \citep{ni2024l2ceval,becker2023programming}.
They also perform well in the reverse direction, namely code summarization, which is directly relevant to reranking methods that use code-to-instruction signals \citep{akib2024analysis}.

\paragraph{Code Reranking.}
A common way to improve code generation is to sample multiple candidate programs and rerank them.
Existing reranking methods can be broadly divided into execution-driven and content-driven approaches.
Execution-driven methods such as CodeT \citep{chen2022codetcodegenerationgenerated} and AgentCoder \citep{huang2024agentcodermultiagentbasedcodegeneration} evaluate candidates by executing them against automatically constructed tests.
Although often effective, such methods depend on the availability and reliability of test suites and may also raise safety concerns when executing untrusted code \citep{yeticstiren2023evaluating,khoury2023secure}.
In this work, we focus on content-driven reranking, where methods use only the original instruction and sampled candidates without executing code.
This setting is more generally applicable when tests are unavailable, unreliable, or unsafe to run.

\paragraph{Coder Reranking.}
A simple and widely used content-driven baseline is to score each candidate independently under the original instruction, i.e., by estimating
\[
\text{P}(c\mid I_0).
\]
Following \citet{chen2021evaluatinglargelanguagemodels}, we refer to this strategy as \emph{Coder} reranking.
Its main limitation is that it scores candidates independently: it does not check code-to-instruction support or compare candidates within the sampled pool.

\paragraph{CoderReviewer Reranking.}
\citet{zhang2022coderreviewerrerankingcode} introduced the idea of augmenting Coder Reranking with a reviewer, which jointly considers how likely a code candidate is under the instruction and how well the instruction is supported by the code.
Formally, the CoderReviewer conditional probability is defined as
\[
\begin{aligned}
\text{P}_{\text{CR}}(c \mid I_0)
&\propto
\text{P}(c \mid I_0) \cdot \text{P}(I_0 \mid c). \\
&\hspace{1.2em}\text{\footnotesize (Coder)}
\hspace{2.6em}
\text{\footnotesize (Reviewer)}
\end{aligned}
\]
By switching the positions of the instruction and code in the conditional formulation, the second term can be interpreted as reformulating the code-generation task as an instruction-generation task.
This bidirectional formulation can be viewed as a specialized form of maximum mutual information \citep{li2016mutualinformationdiversedecoding}.

\paragraph{Consensus-WUCS.}
Consensus-WUCS reranks candidates by weighted inter-candidate agreement \citep{jain-etal-2024-lightweight}. 
For each candidate $c_i$, it computes
\[
\mathrm{GCS}_{\mathrm{WUCS}}(c_i)
=
\frac{1}{n-1}\sum_{j\neq i}\mathrm{WUCS}(c_i,c_j),
\]
where $\mathrm{WUCS}(c_i,c_j)$ measures weighted unigram consistency. 
Its strongest variant combines this consensus score with length-normalized Coder confidence:
\[
\begin{aligned}
\mathrm{score}_{\mathrm{CW}}(c_i)
&=
\mathrm{GCS}_{\mathrm{WUCS}}(c_i) \\
&\quad \cdot
\exp\!\left(
\frac{1}{|c_i|}
\log \mathrm{P}(c_i\mid I_0)
\right).
\end{aligned}
\]
Thus, Consensus-WUCS ranks candidates according to a combination of inter-candidate lexical agreement and length-normalized likelihood under the original instruction.

Overall, these baselines represent direct prompt-conditioned scoring (Coder), bidirectional scoring (CoderReviewer), and inter-candidate consensus (Consensus-WUCS). 
CodeRSA is closest to the latter two, but instead builds a local comparison space from candidate-induced instructions and favors candidates with distinctive, rather than merely central, support.

\section{CodeRSA}
\label{section:CodeRSA}

In this section, we introduce CodeRSA, an RSA-inspired reranking method for code generation.
The key idea is to select not only a candidate that is plausible under the observed instruction \(I_0\), but one that \(I_0\) supports more distinctively than nearby alternatives.

Direct RSA-style inference is difficult because it would require estimating pragmatic distributions over an effectively open-ended space of instructions and programs.
CodeRSA instead builds a finite local comparison space from sampled candidates.
For each candidate \(c_i\), it generates a candidate-induced instruction \(I_i\), a reverse-generated description of the behavior implemented by \(c_i\).
For each pair \((c_i,c_j)\), CodeRSA compares how much more \(I_0\) supports \(c_i\) than \(I_j\) does against how much more \(I_0\) supports \(c_j\) than \(I_i\) does.
Aggregating these pairwise wins gives a tournament-style pragmatic signal, which is combined with a global-support signal for final reranking.

\paragraph{Reverse Generation.}
Given the original instruction \(I_0\), we first sample a pool of candidate programs \(c_1,\dots,c_n\).
For each candidate \(c_i\), we then generate one candidate-induced instruction \(I_i\) using the same base model.
Together, these form the induced instruction set
\[
\mathcal{I}=\{I_0,I_1,\dots,I_n\}.
\]
Intuitively, \(I_i\) represents a candidate-specific interpretation of what a cooperative user might have said if \(c_i\) were the intended answer.

\paragraph{Literal Compatibility.}
As in classic RSA, we begin from a non-pragmatic interpretation layer that evaluates how well a candidate fits an instruction.
In our setting, the base model provides a literal compatibility score for each instruction--candidate pair:
\[
L_0(I_j,c_i):=\log \text{P}(c_i\mid I_j).
\]
Given the induced instruction set \(\mathcal{I}\), we compute this score for every pair \((I_j,c_i)\).
These scores serve as the building blocks for the local pragmatic comparisons below.

\paragraph{Local Pragmatic Contests.}
Starting from the literal compatibility scores \(L_0(I_j,c_i)\), CodeRSA constructs a collection of local pragmatic contests.
Each contest compares two candidates through the smallest contrastive subproblem available in the sampled setting: whether the observed instruction \(I_0\) favors one candidate more strongly than a nearby rival-induced alternative does.
The rationale for this pairwise decomposition is that pragmatic reranking requires contrastive evidence, whereas global competition over the full induced instruction set can be distorted by paraphrastic overlap and by the compressed candidate pool sampled from the same original instruction.
Appendix~\ref{app:theory-rsa-codersa} provides a more detailed motivation.

To make this comparison explicit, consider a restricted binary speaker problem for a fixed candidate \(c_i\), where the available utterances are the observed instruction \(I_0\) and one rival-induced instruction \(I_j\).
Following the RSA intuition that a speaker prefers utterances that better identify the intended meaning, we define
\[
\begin{aligned}
\text{P}_{S_1}(I \mid c_i)
&\propto
\exp\bigl(L_0(I,c_i)\bigr) \cdot \exp(-C(I)), \\
&\qquad I\in\{I_0,I_j\}.
\end{aligned}
\]
where \(C(I)\) is an utterance cost.
Taking the log-odds between \(I_0\) and \(I_j\) gives
\[
\begin{aligned}
\log \frac{\text{P}_{S_1}(I_0\mid c_i)}{\text{P}_{S_1}(I_j\mid c_i)}
&=
L_0(I_0,c_i)-L_0(I_j,c_i) \\
&\qquad -
\bigl(C(I_0)-C(I_j)\bigr).
\end{aligned}
\]
Under the common simplifying assumption that the local cost difference is small in this restricted comparison setting, this motivates the ordered contrast
\[
U_{ij}
=
L_0(I_0,c_i)-L_0(I_j,c_i).
\]
Appendix~\ref{app:theory-rsa-codersa} discusses why omitting explicit cost terms is a reasonable local approximation in our setting.
Thus, \(U_{ij}\) can be interpreted as the cost-free log-odds of a restricted binary speaker comparison between \(I_0\) and \(I_j\) for the fixed candidate \(c_i\).
A large positive value indicates that \(I_0\) identifies \(c_i\) more distinctively than the nearby rival-induced alternative \(I_j\).

A reranker, however, must compare candidates rather than utterances in isolation.
For a pair of candidates \(c_i\) and \(c_j\), we therefore compare the two directed local contests symmetrically:
\[
\begin{aligned}
\Delta_{ij}
&=
\bigl(L_0(I_0,c_i)-L_0(I_j,c_i)\bigr) \\
&\qquad -
\bigl(L_0(I_0,c_j)-L_0(I_i,c_j)\bigr).
\end{aligned}
\]
This margin asks whether \(I_0\) serves as a better distinguishing utterance for \(c_i\) against \(I_j\) than it does for \(c_j\) against \(I_i\).
In this sense, \(\Delta_{ij}\) is the outcome margin of a local pragmatic subgame between \(c_i\) and \(c_j\).
We treat \(c_i\) as winning this subgame if \(\Delta_{ij}>0\), losing if \(\Delta_{ij}<0\), and splitting the comparison evenly in the case of a tie.

Aggregating over all rivals yields the pairwise pragmatic score
\[
\begin{aligned}
\mathrm{pair}_i
&=
\frac{1}{n-1}
\sum_{j\neq i}
\Bigl(
\mathbbm{1}[\Delta_{ij}>0] \\
&\qquad\qquad
+
\frac{1}{2}\mathbbm{1}[\Delta_{ij}=0]
\Bigr).
\end{aligned}
\]
This win-rate aggregation is the normalized Copeland score of candidate \(c_i\) in the tournament induced by the local pragmatic subgames.
We use this tournament perspective rather than summing raw margins because the goal is to measure how consistently a candidate wins local pragmatic comparisons, not to rely on the absolute calibration of a small number of log-probability gaps.
Appendix~\ref{app:pairsearch} compares several alternative aggregation schemes and shows that a tournament-style aggregation is the most stable overall.
Accordingly, \(\mathrm{pair}_i\) should be understood not as a full speaker distribution, but as a tractable tournament aggregation of local RSA-style contests.

\paragraph{Pragmatic Listener.}
In classical RSA, the pragmatic listener combines a speaker signal with a prior:
\[
\text{P}_{L_1}(c \mid I)
\propto
\text{P}_{S_1}(I \mid c)\cdot\text{P}(c).
\]
The speaker term captures how strongly the observed instruction distinguishes a candidate from its alternatives, while the prior favors meanings that remain plausible beyond any single immediate comparison.

CodeRSA preserves this division of labor, but instantiates the second term as a neighborhood-level plausibility measure derived from the induced instruction set rather than as an explicit unconditional prior over candidates:
\[
\mathrm{avg}_i
=
\frac{1}{n+1}\sum_{j=0}^{n}\log \text{P}(c_i\mid I_j).
\]
It measures how broadly candidate \(c_i\) is supported across the induced instruction neighborhood.
In our RSA-inspired formulation, \(\mathrm{avg}_i\) plays the role of a neighborhood prior: rather than estimating an unconditional prior \(P(c)\), it rewards candidates that remain plausible across multiple nearby interpretations of the task.
Equivalently, it functions as a local stability term that prevents the reranker from relying only on pairwise wins against isolated rivals.
Because \(\mathrm{pair}_i\) and \(\mathrm{avg}_i\) live on different raw scales, we standardize them within each task before combining them.

The final reranker then combines the tournament-style pragmatic signal \(\mathrm{pair}_i\) with the neighborhood-prior term \(\mathrm{avg}_i\):
\[
\mathrm{score}(c_i)
=
z(\mathrm{pair}_i)+z(\mathrm{avg}_i),
\]
where \(z(\cdot)\) denotes within-task standardization.
This equal-weight combination is the default setting used throughout the main paper.
Appendix~\ref{app:alpha-search} also explores a weighted variant,
\[
\mathrm{score}_\alpha(c_i)=\alpha\, z(\mathrm{pair}_i)+(1-\alpha)\, z(\mathrm{avg}_i),
\]
and shows that the equal-weight choice is a simple and stable default, while additional tuning of \(\alpha\) can yield further gains in some settings.
The final prediction is
\[
\hat{c}
=
\arg\max_{c_i}\mathrm{score}(c_i).
\]

In this way, the pairwise term provides the main contrastive signal, while the global-support term plays the listener-side role of a prior-like stabilizer.
Together, they favor candidates that are not merely plausible under the original instruction in isolation, but more specifically and more robustly supported by it relative to nearby alternative interpretations.

\section{Experimental Setup}
\label{sec:experiment}

We evaluate whether CodeRSA improves code reranking in a realistic multi-candidate code setting.
Specifically, we compare four reranking methods: Coder, CoderReviewer, Consensus-WUCS, and CodeRSA.
To isolate reranking quality from generation effects, all methods are evaluated on the same candidate pool for each instance.

\subsection{Datasets and Base Models}

We evaluate on HumanEval+, MBPP+, and BigCodeBench.
HumanEval+ and MBPP+ are the EvalPlus versions of HumanEval and MBPP, with substantially more rigorous functional-correctness test suites \citep{chen2021evaluatinglargelanguagemodels,austin2021program,liu2023your}.
BigCodeBench \citep{zhuo2024bigcodebench} is more challenging, with more complex instructions and richer function usage.

These benchmarks represent different reranking regimes.
HumanEval+ and MBPP+ usually provide substantial reranking headroom because correct candidates are often present but not ranked first.
BigCodeBench is more difficult, yielding lower absolute performance for all methods, but provides a useful stress test for reranking robustness.

For each problem, we construct a shared pool of \(n=10\) candidate programs using temperature 1.2.
Appendix~\ref{app:candidate-number} reports a small sanity check on the effect of varying this candidate-pool size.
We apply a lightweight Python AST filter and retain only syntactically valid candidates.
If fewer than 10 valid candidates are obtained, we continue sampling under the same setting until 10 are collected.
Unless otherwise stated, results are averaged over five candidate-generation seeds (42--46), with all methods evaluated on the same candidate pool for each seed.

We use four open-weight models: Llama-3-8B and Llama-3-70B \citep{grattafiori2024llama}, Qwen2.5-7B \citep{qwen25}, and Qwen2.5-32B-Coder \citep{hui2024qwen2}.
All are instruction-following models; Qwen2.5-32B-Coder is code-specialized, while the others are general-purpose instruction-tuned models.
This setup lets us test CodeRSA across model scales and model families.

\subsection{Implementation of Reranking Methods}

\paragraph{Coder.}
Coder scores each candidate \(c_i\) under the original instruction \(I_0\) using the forward conditional probability \(\text{P}(c_i \mid I_0)\) \citep{chen2021evaluatinglargelanguagemodels}.
We use the default benchmark instruction as context and sum log-probabilities over the candidate code tokens only.

\paragraph{CoderReviewer.}
CoderReviewer augments the forward score with a reverse compatibility term \(\text{P}(I_0 \mid c_i)\) \citep{zhang2022coderreviewerrerankingcode}.
We use the same forward prompt as in Coder and a reviewer prompt that asks the model to recover the original instruction from the candidate code.

\paragraph{Consensus-WUCS.}
For Consensus-WUCS, we follow \citet{jain-etal-2024-lightweight}.
Given the shared candidate pool, we compute weighted inter-candidate unigram consistency scores and combine them with a length-normalized Coder confidence term, as described in Section~\ref{sec:background}.
The highest-scoring candidate is selected.

\paragraph{CodeRSA.}
For CodeRSA, we use the same base model for candidate generation, induced-instruction generation, and compatibility scoring unless otherwise stated.
We generate one induced instruction \(I_i\) for each sampled candidate \(c_i\), yielding
\[
\mathcal{I}=\{I_0,I_1,\dots,I_n\}.
\]
Unless otherwise stated, induced instructions are generated by greedy search using two prompt templates.
HumanEval+ and MBPP+ share a concise one-sentence behavior-description prompt, while BigCodeBench uses a more detailed variant that allows multi-sentence behavioral specifications.
Appendix~\ref{app:induced-prompt} provides the prompts, and Appendix~\ref{app:analysis-instruction} reports a manual quality analysis of induced instructions.
The manual analysis suggests that the reverse-generated instructions are generally reliable and specific enough to support the subsequent reranking experiments.

We then compute
$
L_0(I_j,c_i),
$
the sequence log-probability of candidate \(c_i\) under instruction \(I_j\), summed over candidate code tokens only.
We use raw sequence log-likelihoods rather than length-normalized scores, since length normalization can introduce additional biases \citep{zhang2022coderreviewerrerankingcode}.
We derive \(\mathrm{pair}_i\) and \(\mathrm{avg}_i\) from these scores, standardize them within each task, and combine them as
\[
\mathrm{score}(c_i)=z(\mathrm{pair}_i)+z(\mathrm{avg}_i).
\]
The final prediction is the candidate with the highest score.
\begin{table*}[t]
\centering
\small
\renewcommand{\arraystretch}{1.12}
\setlength{\tabcolsep}{4.8pt}
\begin{tabular}{@{}llrrrrr@{}}
\toprule
Dataset & Model & Random & Coder & CoderReviewer & Consensus-WUCS & CodeRSA \\
\midrule
\multirow{4}{*}{HumanEval+}
& Llama-3-8B        & 34.3 & 40.2 & \second{44.5} & 42.9 & \best{47.8} \\
& Qwen2.5-7B        & 49.5 & 49.9 & 52.0 & \second{55.4} & \best{59.9} \\
& Qwen2.5-32B-Coder & 71.2 & 73.5 & \second{75.2} & 74.5 & \best{76.8} \\
& Llama-3-70B     & 47.4 & 52.2 & \second{55.6} & 51.5 & \best{60.4} \\
\midrule
\multirow{4}{*}{MBPP+}
& Llama-3-8B        & 32.6 & 40.3 & 40.5 & \second{41.2} & \best{44.6} \\
& Qwen2.5-7B        & 68.5 & \second{73.8} & \second{73.8} & 72.4 & \best{76.5} \\
& Qwen2.5-32B-Coder & 74.3 & 77.6 & 69.6 & \best{83.7} & \second{83.5} \\
& Llama-3-70B     & 61.1 & 63.9 & 65.0 & \second{70.4} & \best{70.8} \\
\midrule
\multirow{4}{*}{BigCodeBench}
& Llama-3-8B        & 15.1 & 10.1 & 12.8 & \second{17.1} & \best{20.2} \\
& Qwen2.5-7B        & 31.3 & 30.7 & \second{31.9} & 29.7 & \best{34.0} \\
& Qwen2.5-32B-Coder & 41.8 & 45.0 & 44.9 & \best{46.7} & \second{45.4} \\
& Llama-3-70B     & 36.9 & 37.1 & \second{37.6} & 36.1 & \best{39.5} \\
\bottomrule
\end{tabular}
\caption{Main results across all settings. Numbers report mean accuracy over five candidate-generation seeds (42--46). We compare uniform random selection from the final candidate pool (Random), the direct coder likelihood baseline (Coder), bidirectional reranking (CoderReviewer), Consensus-WUCS, and our method CodeRSA, instantiated as \(z(\mathrm{pair}) + z(\mathrm{avg})\). The best result in each row is marked by darker gray shading and boldface; the second-best result is marked by lighter gray shading and underlining. Ties receive the same mark. Full multi-seed significance comparisons are reported in Appendix~\ref{app:significance}.}
\label{tab:main_all_results}
\end{table*}

\begin{table}[t]
\centering
\footnotesize
\begin{tabular*}{\columnwidth}{@{\extracolsep{\fill}}lccc}
\toprule
Model & \textit{pair} & \textit{avg} & CodeRSA \\
\midrule
\multicolumn{4}{l}{\textbf{HumanEval+}} \\
Llama-3-8B        & 40.1 & 32.9 & 47.8 \\
Qwen2.5-7B        & 51.8 & 44.4 & 59.9 \\
Qwen2.5-32B-Coder & 71.1 & 70.4 & 76.8 \\
Llama-3-70B     & 54.9 & 45.4 & 60.4 \\
\midrule
\multicolumn{4}{l}{\textbf{MBPP+}} \\
Llama-3-8B        & 36.2 & 36.8 & 44.6 \\
Qwen2.5-7B        & 68.8 & 72.1 & 76.5 \\
Qwen2.5-32B-Coder & 76.8 & 76.0 & 83.5 \\
Llama-3-70B     & 65.9 & 61.0 & 70.8 \\
\midrule
\multicolumn{4}{l}{\textbf{BigCodeBench}} \\
Llama-3-8B        & 17.9 & 13.3 & 20.2 \\
Qwen2.5-7B        & 33.4 & 30.4 & 34.0   \\
Qwen2.5-32B-Coder & 42.2 & 43.8& 45.4 \\
Llama-3-70B     & 39.0 & 37.1 & 39.5 \\
\bottomrule
\end{tabular*}
\caption{Component analysis of CodeRSA. Numbers report mean reranking accuracy over five candidate-generation seeds (42--46). The column \textit{pair} uses only the pairwise pragmatic comparison component, \textit{avg} uses only the global-support component, and \textit{CodeRSA} uses their combined score.}
\label{tab:component_analysis}
\end{table}
\begin{table}[t]
\centering
\footnotesize
\setlength{\tabcolsep}{4pt}
\renewcommand{\arraystretch}{1.0}
\begin{tabular}{lccc}
\toprule
Model & HEval+ & MBPP+ & BCB \\
\midrule
Llama-3-8B        & 32.9 & 43.7 & 29.5 \\
Qwen2.5-7B        & 32.4 & 21.9 & 27.1 \\
Qwen2.5-32B-Coder & 18.0 & 21.0 & 8.9 \\
Llama-3-70B     & 33.2 & 29.1 & 8.7 \\
\bottomrule
\end{tabular}
\caption{Candidate-pool headroom (Oracle@10 minus Random). HEval+ denotes HumanEval+, and BCB denotes BigCodeBench.}
\label{tab:headroom}
\end{table}

\section{Quantitative Analysis}

Table~\ref{tab:main_all_results} summarizes the main results across all evaluated settings.
We report final selection accuracy for Random, Coder, CoderReviewer, Consensus-WUCS, and CodeRSA, where CodeRSA uses the default score
\(
z(\mathrm{pair}) + z(\mathrm{avg})
\).
For Random, we report the expected accuracy of uniformly sampling one candidate from the final pool, computed as the average fraction of correct candidates per problem.
All reranking methods are evaluated on the same shared candidate pool for each instance, so the comparison isolates reranking quality rather than differences in candidate generation.

Overall, CodeRSA achieves the strongest average result in 10 of the 12 evaluated settings and remains close to the best result in the other two.
Relative to Coder and CoderReviewer, the gains are consistent: CodeRSA outperforms both baselines in all 12 settings.
Compared with Consensus-WUCS, the pattern is more mixed but still favorable overall: CodeRSA is stronger in 10 settings, while Consensus-WUCS is slightly better on MBPP+ and BigCodeBench with Qwen2.5-32B-Coder.

Across datasets, CodeRSA achieves the best result in all four HumanEval+ settings, including a substantial improvement over CoderReviewer on Qwen2.5-7B (59.9 vs.\ 52.0) and Llama-3-70B (60.4 vs.\ 55.6).
On MBPP+, CodeRSA is best for three of the four models and is within 0.2 points of Consensus-WUCS on Qwen2.5-32B-Coder (83.5 vs.\ 83.7).
On BigCodeBench, CodeRSA is also best in three of the four settings; the exception is again Qwen2.5-32B-Coder, where Consensus-WUCS leads by 1.3 points (46.7 vs.\ 45.4).

Table~\ref{tab:headroom} helps contextualize this pattern by reporting candidate-pool headroom, defined as Oracle@10 minus Random.
This quantity gives a simple upper bound on the room available for reranking within the shared sampled pool.
The headroom values show that the available reranking signal varies substantially by both benchmark and model.
BigCodeBench has limited headroom for the stronger Qwen2.5-32B-Coder and Llama-3-70B pools (8.9 and 8.7 points), which helps explain why the strongest rerankers are close in those settings.
For smaller models, however, BigCodeBench still has substantial recoverable signal (29.5 and 27.1 points), so lower absolute accuracy should not be conflated with uniformly low reranking headroom.
More broadly, smaller headroom leaves less room for any reranker to improve selection, whereas larger values indicate that the sampled pool contains more recoverable signal.

The results also vary across model families.
CodeRSA tends to yield larger improvements for smaller instruction-tuned models such as Llama-3-8B and Qwen2.5-7B, while remaining competitive for stronger models such as Llama-3-70B and Qwen2.5-32B-Coder.
A supplementary post-hoc appendix analysis further examines how the relative strengths of rerankers vary with behavioral diversity in the sampled candidate pool (Appendix~\ref{app:diversity_analysis}), and Appendix~\ref{app:significance} reports the full multi-seed significance comparisons.
As an additional robustness check, Appendix~\ref{app:cross_model_reranking} examines a small cross-model reranking setting in which candidate generation and CodeRSA scoring are decoupled.

\paragraph{Component analysis.}
Table~\ref{tab:component_analysis} shows that the full CodeRSA score outperforms either component used in isolation in all 12 settings.
The pairwise component is often the stronger individual signal, especially on HumanEval+ and BigCodeBench, suggesting that local pragmatic comparison provides much of the discriminative effect.
At the same time, \textit{pair} alone is not sufficient.
On MBPP+, the relative strength of the two components is more mixed; for example, \textit{avg} slightly exceeds \textit{pair} on Llama-3-8B (36.8 vs.\ 36.2) and Qwen2.5-7B (72.1 vs.\ 68.8), while \textit{pair} is stronger on the two larger-model settings.
The full method substantially improves over both individual components in these settings, indicating that the two terms are complementary: \textit{pair} sharpens local discrimination, while \textit{avg} stabilizes the ranking across induced instructions.
\paragraph{Comparison with finite vanilla RSA.}
We also compare CodeRSA with a finite vanilla RSA baseline defined on the same sampled candidate and induced instruction sets; the full construction and representative results are reported in Appendix~\ref{app:vanilla-rsa}.
Empirically, this finite RSA baseline remains below CodeRSA on the tested settings, while its comparison with CoderReviewer is mixed.
This pattern is consistent with our theoretical motivation: direct global pragmatic normalization over the induced instruction set appears less well behaved in this reranking regime, whereas the local pairwise design in CodeRSA reduces the impact of paraphrastic redundancy and candidate-specific normalization effects.
The comparison therefore provides additional evidence that controlling these sources of noise and bias is important for making RSA-style reasoning effective in realistic code reranking.

\section{Discussion}
\label{section:Discussion}

CodeRSA contains deliberate simplifications relative to classical RSA\@.
Rather than modeling a full pragmatic speaker over an open-ended instruction space, it approximates pragmatic comparison through local pairwise contests within a candidate-induced neighborhood.
Instead of introducing an explicit prior, it uses the global-support term as a stabilizing signal over nearby induced instructions.
These choices make CodeRSA tractable in realistic code-generation settings while preserving the central RSA intuition that interpretation should depend on contrast with nearby alternatives, not only on isolated plausibility.
Thus, CodeRSA is best understood as a practical approximation to pragmatic reranking rather than a faithful instantiation of classical RSA\@.

A natural direction for future work is to make this pragmatic effort adaptive.
Prior work on human and computational pragmatics suggests that expensive pragmatic reasoning need not be applied uniformly, but can be approximated or invoked selectively depending on task demands \citep{lieder2020resource,tscshantz2023hybrid,dasgupta2018remembrance,pmlr-v235-pu24c}.
Similarly, future reranking systems could allocate contrastive reasoning after sampling, applying stronger pragmatic comparison only when simpler scoring signals are unlikely to resolve the candidate pool.
The post-hoc analysis in Appendix~\ref{app:diversity_analysis} offers a related perspective: CodeRSA appears more useful when candidate solutions are behaviorally diverse, suggesting that pre-reranking proxies for candidate-pool diversity may help decide when contrastive reranking is warranted.

\section{Conclusion}

This paper introduced CodeRSA, an RSA-inspired reranking method for code generation under natural-language uncertainty.
CodeRSA constructs a candidate-induced alternative set around the original instruction, derives a pairwise pragmatic comparison signal and a global-support signal from this local comparison space, and combines them for final reranking.
Across HumanEval+, MBPP+, and BigCodeBench, and across four open-weight instruction-following models, CodeRSA performs most strongly overall.

Our analyses further show that the pairwise component provides the main contrastive signal, while the global-support component contributes a prior-like stabilizing effect.
Taken together, these results suggest that reranking can benefit from a distinct pragmatic perspective: rather than relying only on candidate-wise scoring or consensus-based centrality, CodeRSA asks which candidate is most specifically supported by the original instruction relative to nearby alternatives.

More broadly, this work shows that practical approximations to pragmatic reasoning can be useful in realistic code-generation settings.
We hope it encourages further work on reranking methods that combine tractable inference with linguistically motivated inductive biases, both for code generation and for other forms of language-mediated decision making.

\section{Limitations}
\label{section:Limitations}

A primary limitation of CodeRSA is computational cost.
Because the method scores each sampled candidate under both the original instruction and candidate-induced instructions, its reranking cost grows quadratically with the candidate-pool size.
For this reason, our main experiments use \(n=10\) candidates per instance to keep runtime and hardware usage manageable.
On 100 MBPP+ problems with Qwen2.5-7B, CodeRSA takes 3.16 seconds per instance on a single NVIDIA Tesla A100, excluding candidate generation.
In the same setting, Coder and Consensus-WUCS each take 0.22 seconds per instance, and CoderReviewer takes 0.42 seconds.
Thus, scaling CodeRSA to substantially larger candidate pools would require additional efficiency improvements, and the \(n=10\) setting may limit the diversity of the sampled solution space.
A small sanity check in Appendix~\ref{app:candidate-number} examines this candidate-pool-size trade-off in one representative setting.

At the same time, the overhead can be reduced by applying CodeRSA selectively or in a multi-stage pipeline.
For example, using CoderReviewer to preselect the top five candidates reduces CodeRSA's reranking time to 1.24 seconds per instance.
Appendix~\ref{app:top5-filtering} further shows that such filtering can retain much of the full method's performance, suggesting that practical deployments need not apply the full CodeRSA procedure to every sampled candidate.
More broadly, CodeRSA should be viewed less as a uniformly cost-efficient replacement for simpler rerankers than as evidence for a complementary reranking principle: local contrastive comparison induced by candidate-specific instructions.

Finally, CodeRSA adapts the core contrastive intuition of RSA to the sampled reranking setting, where full inference over an unrestricted instruction space is not tractable.
Its pairwise contests retain a local speaker-like contrastive signal, while the global-support term provides a prior-like stabilizer over the induced neighborhood.
Future work could explore richer cost models, broader alternative spaces, and closer connections between tractable reranking approximations and full RSA-style inference.

\begin{comment}
\begin{table}[t]
\centering
\small
\begin{tabular}{lc}
\toprule
Method & Avg. time \\
\midrule
Coder & 0.22 s \\
CoderReviewer & 0.42 s \\
Consensus-WUCS & 0.22 s \\
CodeRSA ($n{=}10$) & 3.16 s \\
CodeRSA (top-5 filtered with CoderReviewer) & 1.24 s \\
\bottomrule
\end{tabular}
\caption{Average reranking time per instance on 100 MBPP+ problems using Qwen2.5-7B, measured as wall-clock time on a single NVIDIA Tesla A100 (PCIe 4.0, 80 GB HBM2e, 300 W) with batch size 10. Candidate generation is excluded.}
\label{tab:efficiency}
\end{table}
\end{comment}

\bibliography{custom}

\clearpage
\appendix

\section{Appendix}

\subsection{Motivating the Pairwise Design}
\label{app:theory-rsa-codersa}

This appendix explains why the pairwise design used in CodeRSA is a natural approximation to the contrastive role of RSA in realistic code reranking.
Our goal is not to instantiate a full pragmatic speaker over an unrestricted instruction space, but to clarify what vanilla RSA would look like in the current sampled setting and why that full competition becomes problematic.

\paragraph{Vanilla RSA in the sampled setting.}
Suppose we restrict attention to the sampled candidate set
\[
\mathcal{C}=\{c_1,\dots,c_n\}
\]
and the induced instruction set
\[
\mathcal{I}=\{I_0,I_1,\dots,I_n\}.
\]
In this finite setting, the literal listener evaluates candidates under each instruction:
\[
P_{L_0}(c_i\mid I)
=
\frac{P(c_i\mid I)}{\sum_{k=1}^{n}P(c_k\mid I)}.
\]
The pragmatic speaker then compares all instructions in \(\mathcal I\) for each candidate.
With a cost function \(C(I)\), this gives
\[
\begin{aligned}
P_{S_1}(I_j\mid c_i)
&=
\frac{P_{L_0}(c_i\mid I_j)\exp(-C(I_j))}{Z_i},\\
Z_i
&=
\sum_{m=0}^{n}P_{L_0}(c_i\mid I_m)\exp(-C(I_m)).
\end{aligned}
\]
If we ignore speaker costs for the present discussion, this reduces to
\[
P_{S_1}(I_j\mid c_i)
=
\frac{P_{L_0}(c_i\mid I_j)}
{\sum_{m=0}^{n}P_{L_0}(c_i\mid I_m)}.
\]
Finally, the pragmatic listener ranks candidates under the observed instruction \(I_0\) according to
\[
P_{L_1}(c_i\mid I_0)\propto P_{S_1}(I_0\mid c_i)P(c_i).
\]

Under this finite approximation, vanilla RSA therefore selects candidates by asking whether the original instruction \(I_0\) remains speaker-preferred for that candidate relative to the full induced instruction set.
The difficulty is that this global normalization is precisely where the sampled code-generation setting introduces systematic distortions.

\paragraph{Problem 1: paraphrastic dilution.}
A direct global competition over induced instructions is vulnerable to paraphrastic redundancy.
Suppose a set of alternatives \(\mathcal{P}=\{I_{m_1},\dots,I_{m_r}\}\) are near-equivalent for candidate \(c_i\), in the sense that
\[
\begin{aligned}
P_{L_0}(c_i\mid I_{m_1})
&\approx \cdots \approx
P_{L_0}(c_i\mid I_{m_r}) \\
&\approx p_i.
\end{aligned}
\]
Then their total contribution to a speaker-style normalization is approximately
\[
\begin{aligned}
&\sum_{I_m\in\mathcal P}
P_{L_0}(c_i\mid I_m)\exp(-C(I_m)) \\
&\qquad\approx
r\,p_i\,\bar w_C,
\end{aligned}
\]
where \(\bar w_C\) is an average cost weight.
The crucial point is that this term grows with the \emph{number} of paraphrastic alternatives, even when they do not introduce genuinely new semantic competition.
As a result, full competition over the induced instruction set can spend contrastive capacity separating formulations rather than separating candidate behaviors.

The pairwise design weakens this effect by replacing global accumulation with local comparison.
In vanilla RSA, all paraphrastic alternatives contribute jointly to a single normalization term, so their effect grows directly with their number.
By contrast, the local quantity
\[
U_{ij}
=
L_0(I_0,c_i)-L_0(I_j,c_i)
\]
compares \(I_0\) with one nearby alternative at a time.
If \(I_j\) is only a paraphrastic variant of \(I_0\), then this difference is typically small, so such alternatives act as weak local competitors rather than jointly inflating a global competition term.
This effect is further reduced by the final tournament-style win-rate aggregation, which tracks only whether a candidate repeatedly wins local contrasts rather than summing the magnitudes of all local margins.
In this sense, the pairwise design alleviates paraphrastic dilution by converting a global multiplicity effect into a collection of weak local tests.

\paragraph{Problem 2: compression induced by sampling under \(I_0\).}
A second distortion arises on the candidate side.
The candidate pool \(\mathcal{C}=\{c_1,\dots,c_n\}\) is sampled conditioned on the original instruction \(I_0\), so many sampled candidates already receive relatively large literal-listener scores under \(I_0\):
\[
\begin{gathered}
P(c_i\mid I_0)
\text{ is often non-negligible}\\
\text{for many } c_i\in\mathcal{C}.
\end{gathered}
\]
Equivalently, the \(I_0\)-conditioned row of literal-listener scores often has the form
\[
L_0(I_0,c_i)=B+\varepsilon_i,
\]
where \(B\) is a shared high baseline across candidates and \(\varepsilon_i\) contains the remaining candidate-specific variation.
In this regime, global competition depends on quantities whose absolute scale is already uniformly biased upward, so the relative differences needed for discrimination are compressed.

Here again, the local difference \(U_{ij}\) is helpful.
Because it is defined as a difference between the support of \(c_i\) under \(I_0\) and under one nearby alternative \(I_j\), it is less sensitive to the absolute scale of \(L_0(I_0,c_i)\) than a full global comparison would be.
The effect becomes even clearer after symmetrization.
CodeRSA defines
\[
\Delta_{ij}=U_{ij}-U_{ji},
\]
which asks whether \(I_0\) more strongly singles out \(c_i\) against \(c_j\) than it singles out \(c_j\) against \(c_i\).
Substituting \(L_0(I_0,c_k)=B+\varepsilon_k\) gives
\[
\begin{aligned}
\Delta_{ij}
&=
[(B+\varepsilon_i)-L_0(I_j,c_i)]\\
&\quad -
[(B+\varepsilon_j)-L_0(I_i,c_j)],
\end{aligned}
\]
so that the shared baseline \(B\) cancels.
Thus, the symmetric pairwise margin compares candidates using relative local differences rather than absolute \(I_0\)-conditioned support.
This is precisely the form of comparison needed to counteract compression induced by sampling under \(I_0\).

\paragraph{Why omitting local cost terms is reasonable here.}
The pairwise construction in CodeRSA ultimately compares candidates through the symmetric margin
\[
\Delta_{ij}=U_{ij}-U_{ji}.
\]
To see what cost approximation is actually involved, start from the cost-sensitive binary speaker contrasts
\[
\begin{aligned}
U^{\mathrm{full}}_{ij}
&=
L_0(I_0,c_i)-L_0(I_j,c_i)\\
&\quad-\bigl(C(I_0)-C(I_j)\bigr),
\end{aligned}
\]
and
\[
\begin{aligned}
U^{\mathrm{full}}_{ji}
&=
L_0(I_0,c_j)-L_0(I_i,c_j)\\
&\quad-\bigl(C(I_0)-C(I_i)\bigr).
\end{aligned}
\]
Their symmetric difference is
\[
\begin{aligned}
\Delta^{\mathrm{full}}_{ij}
&=
U^{\mathrm{full}}_{ij}-U^{\mathrm{full}}_{ji} \\
&=
\bigl[L_0(I_0,c_i)-L_0(I_j,c_i)\bigr] \\
&\quad -
\bigl[L_0(I_0,c_j)-L_0(I_i,c_j)\bigr] \\
&\qquad +
\bigl(C(I_j)-C(I_i)\bigr).
\end{aligned}
\]
Thus, the shared cost of the observed instruction \(I_0\) cancels exactly in the final symmetric comparison.
The only remaining cost term is the residual difference between the two rival-induced instructions, \(C(I_j)-C(I_i)\).

This residual approximation is more limited than dropping all utterance costs in a full RSA model.
In CodeRSA, the competing instructions \(I_i\) and \(I_j\) are not arbitrary utterances drawn from an open-ended instruction space.
They are generated from the same sampled candidate pool, using the same reverse-generation procedure, for the same underlying task.
As a result, they are procedurally homogeneous and semantically local: although they may still differ in length, specificity, or syntactic form, they are all alternative descriptions of nearby candidate programs within one shared comparison space.

For this reason, we treat the residual cost difference \(C(I_j)-C(I_i)\) as secondary to the compatibility differences captured by the corresponding \(L_0\) terms.
Equivalently, the pairwise term is intended to approximate local pragmatic comparison within a shared candidate-induced neighborhood, rather than to model the full pragmatic speaker exactly.
Under this interpretation, omitting explicit cost terms is a reasonable simplification for within-pool comparison, even though richer cost models remain a natural direction for future work.

\begin{table*}[t]
\centering
\small
\begin{tabular*}{\textwidth}{@{\extracolsep{\fill}}llccc@{}}
\toprule
Dataset & Model & Win rate (main) & Mean margin & Positive-margin sum \\
\midrule
HumanEval+ & Llama-3-8B & 40.1 & 38.2 & 37.4 \\
HumanEval+ & Qwen2.5-7B & 51.8 & 50.4 & 50.0 \\
MBPP+ & Llama-3-8B & 36.2 & 35.1 & 35.4 \\
MBPP+ & Qwen2.5-7B & 68.8 & 67.8 & 67.1 \\
\bottomrule
\end{tabular*}
\caption{Comparison of alternative aggregation schemes for the pairwise component on four representative settings. Numbers report mean reranking accuracy over five candidate-generation seeds (42--46). ``Win rate'' is the tournament-style aggregation used in the main paper.}
\label{tab:pairsearch}
\end{table*}
\subsection{Alternative Pairwise Aggregation Schemes}
\label{app:pairsearch}

The pairwise component of CodeRSA is implemented in the main paper as a tournament-style win rate over pairwise margins.
This choice is motivated by robustness: the goal of the pairwise signal is to capture whether the original instruction \(I_0\) consistently singles out the same candidate across nearby alternatives, rather than to rely on the raw magnitude of a small number of pairwise comparisons.

To validate this design choice, we compare three candidate-level aggregation schemes on four representative settings: HumanEval+ and MBPP+, each with Llama-3-8B and Qwen2.5-7B.
These settings focus on smaller open-weight models and the two smaller benchmarks, where reranking headroom is substantial and the candidate-induced comparison space is large enough to test aggregation behavior.
All results in Table~\ref{tab:pairsearch} report mean reranking accuracy over the same five candidate-generation seeds (42--46) used in the main experiments.
All variants start from the same pairwise preference margin
\[
\begin{aligned}
\Delta_{ij}
&=
\bigl[L_0(I_0,c_i)-L_0(I_j,c_i)\bigr]\\
&\quad -
\bigl[L_0(I_0,c_j)-L_0(I_i,c_j)\bigr],
\end{aligned}
\]
which compares whether the original instruction \(I_0\) supports candidate \(c_i\) over candidate \(c_j\) more strongly than it supports \(c_j\) over \(c_i\), once each candidate is contrasted against the other's induced interpretation.

The first aggregation scheme is the \emph{tournament-style win rate} used in the main paper:
\[
\mathrm{pair}^{\mathrm{win}}_i
=
\frac{1}{n-1}\sum_{j\neq i}
\begin{cases}
1, & \Delta_{ij}>0,\\
0.5, & \Delta_{ij}=0,\\
0, & \Delta_{ij}<0.
\end{cases}
\]
This formulation measures how often candidate \(c_i\) is preferred in pairwise comparison against the rest of the sampled candidate set.
Its main advantage is that it emphasizes consistency across local contrasts while reducing the influence of a small number of unusually large margins.

The second scheme is the \emph{mean margin}:
\[
\mathrm{pair}^{\mathrm{mean}}_i
=
\frac{1}{n-1}\sum_{j\neq i}\Delta_{ij}.
\]
This alternative preserves the signed magnitude of each pairwise comparison rather than reducing it to a discrete win or loss.
It is therefore a more direct way of aggregating pairwise evidence.
However, it also assumes that the numerical scale of \(\Delta_{ij}\) is comparable across different induced instructions, which may not hold reliably in practice because these values can be affected by wording, length, and local likelihood calibration.

The third scheme is the \emph{positive-margin sum}:
\[
\mathrm{pair}^{+}_i
=
\sum_{j\neq i}\max(\Delta_{ij},0).
\]
This variant rewards candidates for strong positive margins while ignoring losses except indirectly through the absence of reward.
Compared with mean margin, it is less sensitive to large negative outliers, but it still allows a small number of large positive comparisons to dominate the final score.

Table~\ref{tab:pairsearch} compares these three aggregation schemes.
The results support the tournament-style win-rate formulation across all four tested settings.
Win rate outperforms both raw-margin alternatives on HumanEval+ with Llama-3-8B (40.1 vs.\ 38.2 and 37.4), HumanEval+ with Qwen2.5-7B (51.8 vs.\ 50.4 and 50.0), MBPP+ with Llama-3-8B (36.2 vs.\ 35.1 and 35.4), and MBPP+ with Qwen2.5-7B (68.8 vs.\ 67.8 and 67.1).
The differences are modest in some settings, but the direction is consistent.

Overall, these results indicate that aggregation based directly on raw margin values does not provide a more reliable pairwise signal in these representative small-model settings.
By contrast, the tournament-style formulation remains the strongest of the tested variants and is especially appropriate when local contrast needs to be aggregated conservatively.
This pattern supports our choice of win-rate aggregation as a stable, noise-reducing approximation to the contrastive role of the RSA speaker.

\subsection{A Pragmatic Knob for Balancing Local Contrast and Global Support}
\label{app:alpha-search}

\begin{table*}[ht]
\centering
\small
\begin{tabular*}{\textwidth}{@{\extracolsep{\fill}}llcccc@{}}
\toprule
Dataset & Model & Equal weight (\(\alpha=0.5\)) & Best \(\alpha\) & Best accuracy & Gain \\
\midrule
HumanEval+ & Llama-3-8B & 47.80 & 0.50 & 47.80 & +0.00 \\
HumanEval+ & Qwen2.5-7B & 59.88 & 0.48 & 60.12 & +0.24 \\
MBPP+ & Llama-3-8B & 44.55 & 0.45 & 44.81 & +0.26 \\
MBPP+ & Qwen2.5-7B & 76.51 & 0.50 & 76.51 & +0.00 \\
\bottomrule
\end{tabular*}
\caption{Effect of reweighting the pairwise and global-support components on four representative settings. Results report mean accuracy over the five candidate-generation seeds (42--46). The best \(\alpha\) remains in the interior of the range in all four cases, indicating that both components contribute, while moderate tuning of this pragmatic knob can yield further gains over the equal-weight default.}
\label{tab:alpha-search}
\end{table*}
In the main paper, we use the equal-weight combination
\[
\mathrm{score}(c_i)=z(\mathrm{pair}_i)+z(\mathrm{avg}_i)
\]
as a simple tuning-free default.
More generally, however, the relative weighting between the pairwise and global-support terms can be interpreted as a pragmatic knob: it controls how strongly the reranker prefers local pragmatic discrimination over broader support across induced instructions.

Specifically, we consider the weighted variant
\[
\mathrm{score}_\alpha(c_i)=\alpha z(\mathrm{pair}_i)+(1-\alpha)z(\mathrm{avg}_i),
\]
where \(\alpha\in[0,1]\).
Here, \(\alpha=1\) corresponds to pairwise-only reranking, \(\alpha=0\) corresponds to global-support-only reranking, and intermediate values interpolate between the two.
Under this interpretation, larger \(\alpha\) places more emphasis on local pragmatic contrast, while smaller \(\alpha\) places more emphasis on broader support across the induced instruction set.

We evaluate a small grid of mixture weights for \(\alpha\).
To keep this analysis lightweight while covering different settings, we report four representative cases: HumanEval+ and MBPP+, each with Llama-3-8B and Qwen2.5-7B.
Results are averaged over the same five candidate-generation seeds (42--46) used in the main experiments.

Table~\ref{tab:alpha-search} summarizes the results.
The best-performing values of \(\alpha\) remain close to the equal-weight default, ranging from 0.45 to 0.50.
This suggests that, in these representative small-model settings, the default balance between local contrast and global support is already near the best point in the searched range.
At the same time, the best values remain in the interior of the range rather than collapsing to either \(\alpha=0\) or \(\alpha=1\), which reinforces the main conclusion that both components contribute meaningfully.
The equal-weight setting used in the main paper should therefore be viewed as a simple and stable default, with only modest additional gains from tuning this pragmatic knob in these settings.
\begin{figure*}[t]
\centering
\includegraphics[width=0.70\textwidth]{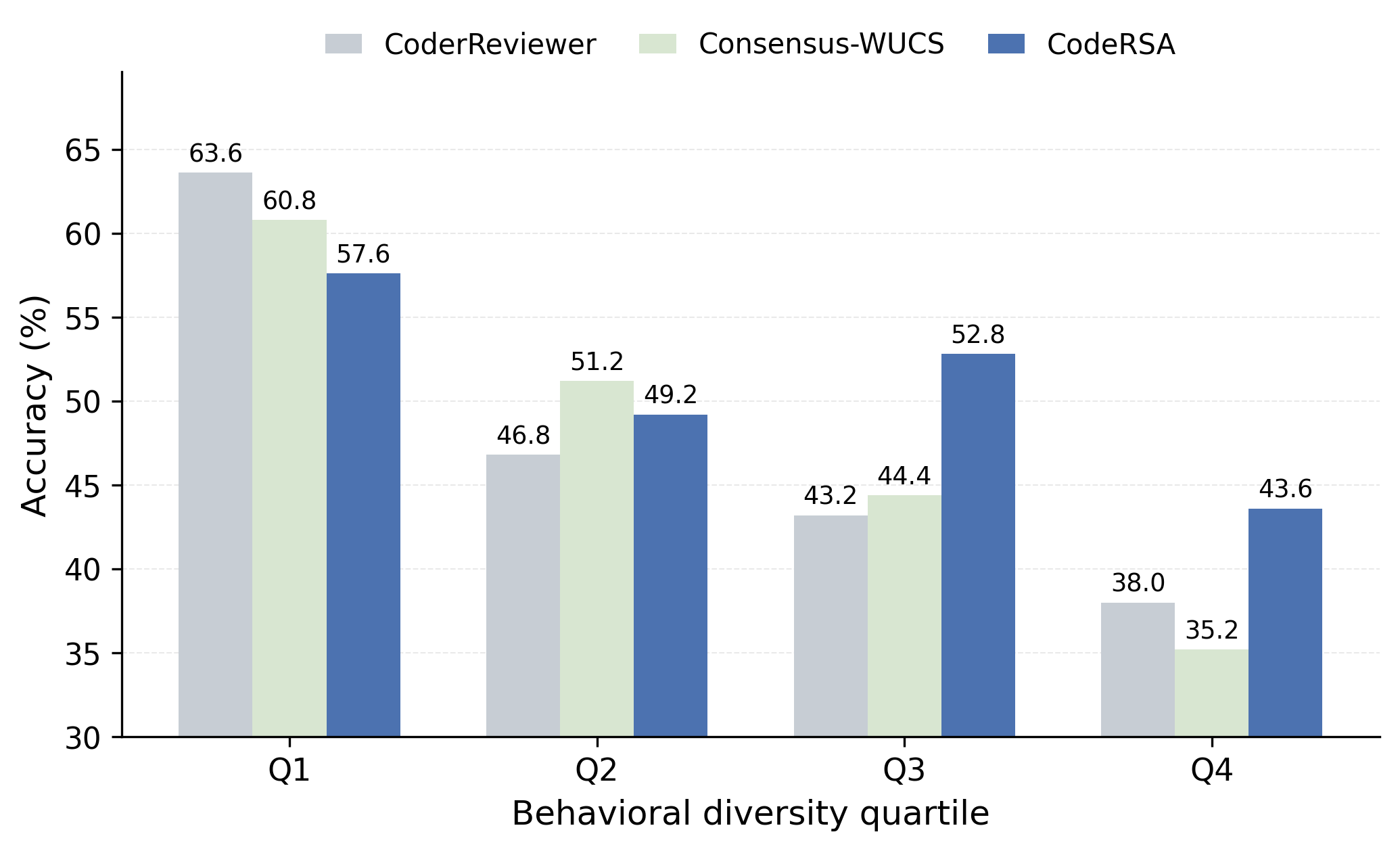}
\caption{
    Accuracy across quartiles of behavioral diversity, from lowest (Q1) to highest (Q4), on the pooled post-hoc sample described in this appendix.
    CoderReviewer and Consensus-WUCS are strongest in the lowest-diversity quartile, Consensus-WUCS remains strongest in Q2, and CodeRSA is strongest in the two higher-diversity quartiles.
    These patterns should be interpreted descriptively because the quartiles pool across datasets, models, and seeds.
}
\label{fig:diversity_accuracy}
\end{figure*}

\subsection{Post-hoc Analysis of Behavioral Diversity and Reranking Performance}
\label{app:diversity_analysis}

This appendix provides an additional post-hoc analysis of the association between behavioral diversity in the sampled candidate pool and reranking performance.
The diversity statistic is computed only after candidate generation and evaluation, and is used solely to stratify instances for analysis.
It is not part of the reranking procedure and is not assumed to be available at inference time.

For each problem, we begin from the same shared candidate pool of \(n=10\) programs used in the main experiments.
We then execute these candidates on the available benchmark test cases and measure \emph{behavioral diversity} as the average pairwise disagreement in their observed test outcomes.
Formally, for a candidate pool \(C=\{c_1,\dots,c_n\}\), we define
\[
\begin{aligned}
\mathrm{Div}(C)
&=
\frac{2}{n(n-1)}
\sum_{i<j}
\frac{1}{T}
\sum_{t=1}^{T}
\mathbf{1}\!\left[
    b_i^{(t)} \neq b_j^{(t)}
\right].
\end{aligned}
\]

where \(b_i^{(t)}\) denotes the observed outcome of candidate \(c_i\) on test case \(t\), and \(T\) is the number of available test cases for that problem.
Intuitively, \(\mathrm{Div}(C)\) is high when candidates behave differently under execution and low when they behave similarly, regardless of surface-form differences in the code.

To obtain a broad sample for analysis, we randomly draw 200 seed-specific instances from each of the five candidate-generation seeds used in the main experiments, yielding 1000 instances in total.
This sampling is performed jointly over all dataset and model combinations rather than separately within each benchmark or model.
As a result, the strata below should be interpreted as descriptive groups over a mixed pool of instances rather than as controlled comparisons within a fixed dataset-model setting.

For each sampled instance, we compute \(\mathrm{Div}(C)\), rank all sampled instances by this value, and partition them into four quartiles from lowest to highest behavioral diversity.
We then evaluate each reranking method separately within each quartile.
Figure~\ref{fig:diversity_accuracy} reports the resulting accuracies; each quartile contains 250 sampled instances.

Figure~\ref{fig:diversity_accuracy} shows that the relative ordering of rerankers varies across the pooled diversity quartiles.
CoderReviewer and Consensus-WUCS are strongest in the lowest-diversity quartile, Consensus-WUCS remains strongest in Q2, and CodeRSA is strongest in Q3 and Q4.
This pattern is consistent with the intuition that local contrastive reranking becomes more useful when the sampled candidate pool contains more behaviorally distinct alternatives.

At the same time, this analysis is descriptive rather than causal.
Because the quartiles are formed over a pooled sample spanning different datasets, models, and seeds, behavioral diversity is not isolated from other factors such as task difficulty, model strength, and candidate-pool quality.
Accordingly, the figure should be read as a suggestive empirical pattern rather than as a controlled estimate of the independent effect of diversity on reranking performance.
\begin{table*}[!t]
\centering
\small

\begin{subtable}{\textwidth}
\centering
\begin{tabular*}{\textwidth}{@{\extracolsep{\fill}}llcccc@{}}
\toprule
Dataset & Model & Coder & CoderReviewer & Consensus-WUCS & CodeRSA \\
\midrule
\multirow{4}{*}{HumanEval+}
& Llama-3-8B & 40.2 $\pm$ 2.8 & 44.5 $\pm$ 2.0 & 42.9 $\pm$ 1.5 & 47.8 $\pm$ 2.2 \\
& Qwen2.5-7B & 49.9 $\pm$ 2.7 & 52.0 $\pm$ 0.8 & 55.4 $\pm$ 2.1 & 59.9 $\pm$ 1.4 \\
& Qwen2.5-32B-Coder & 73.5 $\pm$ 0.7 & 75.2 $\pm$ 2.3 & 74.5 $\pm$ 0.5 & 76.8 $\pm$ 2.1 \\
& Llama-3-70B & 52.2 $\pm$ 2.7 & 55.6 $\pm$ 2.7 & 51.5 $\pm$ 1.7 & 60.4 $\pm$ 2.9 \\
\midrule
\multirow{4}{*}{MBPP+}
& Llama-3-8B & 40.3 $\pm$ 2.0 & 40.5 $\pm$ 2.2 & 41.2 $\pm$ 0.7 & 44.6 $\pm$ 1.0 \\
& Qwen2.5-7B & 73.8 $\pm$ 2.5 & 73.8 $\pm$ 2.2 & 72.4 $\pm$ 0.9 & 76.5 $\pm$ 2.6 \\
& Qwen2.5-32B-Coder & 77.6 $\pm$ 2.0 & 69.6 $\pm$ 1.0 & 83.7 $\pm$ 0.3 & 83.5 $\pm$ 0.7 \\
& Llama-3-70B & 63.9 $\pm$ 3.0 & 65.0 $\pm$ 1.1 & 70.4 $\pm$ 0.9 & 70.8 $\pm$ 1.0 \\
\midrule
\multirow{4}{*}{BigCodeBench}
& Llama-3-8B & 10.1 $\pm$ 1.4 & 12.8 $\pm$ 1.4 & 17.1 $\pm$ 0.8 & 20.2 $\pm$ 0.7 \\
& Qwen2.5-7B & 30.7 $\pm$ 0.6 & 31.9 $\pm$ 0.9 & 29.7 $\pm$ 1.8 & 34.0 $\pm$ 0.6 \\
& Qwen2.5-32B-Coder & 45.0 $\pm$ 0.8 & 44.9 $\pm$ 0.9 & 46.7 $\pm$ 0.6 & 45.4 $\pm$ 0.4 \\
& Llama-3-70B & 37.1 $\pm$ 0.4 & 37.6 $\pm$ 0.7 & 36.1 $\pm$ 0.5 & 39.5 $\pm$ 0.5 \\
\bottomrule
\end{tabular*}
\caption{Mean accuracy and standard deviation over five candidate-generation seeds (42--46) for the three main baselines and CodeRSA. For each seed, all reranking methods are evaluated on the same shared candidate pool.}
\label{tab:multi_seed_results}
\end{subtable}

\vspace{1em}

\begin{subtable}{\textwidth}
\centering
\begin{tabular*}{\textwidth}{@{\extracolsep{\fill}}lllcc@{}}
\toprule
Dataset & Model & Baseline & $\Delta$ (mean) & Significant seeds \\
\midrule
\multirow{8}{*}{HumanEval+}
& Llama-3-8B         & CR    & +3.3  & 2/5 \\
&                     & CWUCS & +4.9  & 3/5 \\
& Qwen2.5-7B          & CR    & +7.9  & 3/5 \\
&                     & CWUCS & +4.5  & 4/5 \\
& Qwen2.5-32B-Coder   & CR    & +1.6  & 0/5 \\
&                     & CWUCS & +2.3  & 2/5 \\
& Llama-3-70B       & CR    & +4.8  & 2/5 \\
&                     & CWUCS & +8.9  & 5/5 \\
\midrule
\multirow{8}{*}{MBPP+}
& Llama-3-8B          & CR    & +4.1  & 2/5 \\
&                     & CWUCS & +3.4  & 2/5 \\
& Qwen2.5-7B          & CR    & +2.7  & 0/5 \\
&                     & CWUCS & +4.1  & 4/5 \\
& Qwen2.5-32B-Coder   & CR    & +13.9 & 5/5 \\
&                     & CWUCS & -0.2  & 0/5 \\
& Llama-3-70B       & CR    & +5.8  & 4/5 \\
&                     & CWUCS & +0.4  & 0/5 \\
\midrule
\multirow{8}{*}{BigCodeBench}
& Llama-3-8B          & CR    & +7.4  & 5/5 \\
&                     & CWUCS & +3.1  & 2/5 \\
& Qwen2.5-7B          & CR    & +2.1  & 2/5 \\
&                     & CWUCS & +4.3  & 4/5 \\
& Qwen2.5-32B-Coder   & CR    & +0.5  & 0/5 \\
&                     & CWUCS & -1.3  & 0/5 \\
& Llama-3-70B       & CR    & +1.9  & 2/5 \\
&                     & CWUCS & +3.4  & 3/5 \\
\bottomrule
\end{tabular*}
\caption{Multi-seed summary of paired significance results over five candidate-generation seeds (42--46). $\Delta$ reports the mean accuracy difference of CodeRSA relative to the corresponding baseline. Significant seeds reports how many of the five seeds yield a significant advantage for CodeRSA under an exact paired McNemar test (\(p<0.05\)).}
\label{tab:significance_summary}
\end{subtable}

\caption{Additional multi-seed results and paired significance analysis.}
\label{tab:multi_seed_combined}
\end{table*}

\subsection{Additional Multi-Seed Analysis}
\label{app:significance}
To further assess the robustness of CodeRSA to candidate-pool variation, we reran all evaluated settings with five candidate-generation seeds (42--46).
For each seed, all reranking methods operate on the same shared candidate pool, so the resulting variation reflects sensitivity to candidate sampling rather than differences in generation across methods.

Table~\ref{tab:multi_seed_results} reports mean accuracy and standard deviation over these five seeds for the three main baselines and CodeRSA.
Overall, the multi-seed results are consistent with the main-table picture: CodeRSA is usually strongest, but the size of its advantage depends on the dataset, model, and baseline.
The gains over CoderReviewer are broad, appearing in all evaluated settings, with especially large improvements on MBPP+ with Qwen2.5-32B-Coder and on BigCodeBench with Llama-3-8B.
Compared with Consensus-WUCS, the pattern is more mixed.
CodeRSA is clearly stronger on several HumanEval+ and BigCodeBench settings, but Consensus-WUCS remains competitive or stronger in some cases, such as MBPP+ with Qwen2.5-32B-Coder and BigCodeBench with Qwen2.5-32B-Coder.
Thus, the multi-seed results support the overall robustness of CodeRSA while also showing that its advantage over consensus-based reranking is not uniform across regimes.
\begin{table*}[t]
\centering
\small
\begin{tabular*}{\textwidth}{@{\extracolsep{\fill}}llcc}
\toprule
Dataset & Candidate model & \multicolumn{2}{c}{CodeRSA reranker model} \\
\cmidrule(lr){3-4}
 & & Llama-3-8B & Qwen2.5-7B \\
\midrule
\multirow{2}{*}{HumanEval+}
& Llama-3-8B  & 47.8 & 48.7 (+0.9) \\
& Qwen2.5-7B  & 58.4 (-1.5) & 59.9 \\
\midrule
\multirow{2}{*}{MBPP+}
& Llama-3-8B  & 44.6 & 45.8 (+1.2) \\
& Qwen2.5-7B  & 76.9 (+0.4) & 76.5 \\
\bottomrule
\end{tabular*}
\caption{Cross-model reranking sanity check. Candidate model denotes the model used to generate the candidate pool; reranker model denotes the model used to generate induced instructions and compute CodeRSA scores. Diagonal entries are the standard same-model setting. Parentheses report the change relative to the same-model CodeRSA result for the same dataset and candidate model.}
\label{tab:cross_model_reranking}
\end{table*}
Table~\ref{tab:significance_summary} complements this view by summarizing paired significance across the same five seeds.
For each setting, we compare CodeRSA against CoderReviewer and Consensus-WUCS using an exact paired McNemar test on each seed separately, and report how many of the five seeds yield a significant advantage for CodeRSA.
This provides a robustness-oriented view of significance: rather than asking only whether one fixed candidate pool yields a significant gain, it asks how often the gain remains significant across independently sampled candidate pools.

Several patterns emerge.
First, CodeRSA's gains over CoderReviewer are the most stable where the mean improvements are largest.
For example, CodeRSA is significant against CoderReviewer in all five seeds on MBPP+ with Qwen2.5-32B-Coder and on BigCodeBench with Llama-3-8B, and in four of five seeds on MBPP+ with Llama-3-70B.
Second, HumanEval+ shows a more moderate but still positive pattern: the strongest significance evidence appears for Qwen2.5-7B and Llama-3-70B, while smaller mean gains yield fewer significant seeds.
Third, comparisons with Consensus-WUCS are less uniform.
In several settings CodeRSA improves over CWUCS by multiple points, but in others the mean difference is small or negative, reflecting the fact that consensus-based reranking remains a strong competitor in some candidate pools.

Taken together, Tables~\ref{tab:multi_seed_results} and~\ref{tab:significance_summary} support a more nuanced interpretation of the main results.
CodeRSA is strongest overall, and its largest improvements over CoderReviewer are robust across candidate-generation seeds.
At the same time, not every positive average gain should be interpreted equally strongly: when the mean difference is small, or when Consensus-WUCS is already very competitive, the seed-level evidence is correspondingly weaker.
This pattern is consistent with the broader empirical picture of the paper: CodeRSA is most beneficial when local pragmatic contrast provides information not already captured by candidate-wise likelihood or consensus, while gains become smaller and less decisive when another reranker already captures much of the recoverable signal.

\subsection{Cross-Model Reranking Sanity Check}
\label{app:cross_model_reranking}

In the main experiments, CodeRSA uses the same base model to generate the candidate pool, generate candidate-induced instructions, and compute reranking scores. This design is natural because it keeps generation and scoring within a single model distribution. However, it also raises a question about whether the reranking signal depends strongly on using the same model for both candidate generation and CodeRSA scoring.

To examine this, we conduct a small cross-model sanity check on HumanEval+ and MBPP+ using Llama-3-8B and Qwen2.5-7B. For each dataset, we fix the candidate pool generated by one model, but allow the CodeRSA reranker model to vary. The reranker model is responsible for generating candidate-induced instructions and computing the literal compatibility scores used by CodeRSA. The diagonal entries in Table~\ref{tab:cross_model_reranking} correspond to the standard same-model setting used in the main experiments, while the off-diagonal entries evaluate cross-model reranking.

The results suggest that CodeRSA is not highly sensitive to using the same model for candidate generation and reranking. On HumanEval+, using Qwen2.5-7B as the reranker for Llama-3-8B candidates slightly improves accuracy from 47.8 to 48.7, while using Llama-3-8B as the reranker for Qwen2.5-7B candidates decreases accuracy from 59.9 to 58.4. On MBPP+, the cross-model reranker slightly improves over the same-model setting in both directions, from 44.6 to 45.8 for Llama-3-8B candidates and from 76.5 to 76.9 for Qwen2.5-7B candidates. These small changes indicate that the local pragmatic comparison signal can transfer across nearby instruction-following models, rather than merely exploiting model-specific likelihood artifacts.

At the same time, this analysis is limited in scope. It covers only two models, two smaller benchmarks, and the default candidate-pool size. We therefore treat it as a sanity check rather than as evidence for a general cross-model reranking strategy. More systematic study of heterogeneous candidate generation and pragmatic reranking remains a useful direction for future work.
\begin{table*}[!t]
\centering
\footnotesize
\renewcommand{\arraystretch}{1.3}
\setlength{\tabcolsep}{4pt}

\begin{tabular}{@{}p{0.14\textwidth} p{0.40\textwidth} p{0.40\textwidth}@{}}
\toprule
& \textbf{Coder/CR-selected $c_4$} & \textbf{CodeRSA-selected $c_7$} \\
\midrule

\textbf{Task}
& \multicolumn{2}{p{0.80\textwidth}@{}}{Write a function to find the product of the first even and odd number in a given list.} \\

\textbf{Passes tests}
& No
& Yes \\

\textbf{Scores}
& \makecell[l]{Coder: $-23.06$ \\ CR: $-69.93$}
& \makecell[l]{Coder: $-36.67$ \\ CR: $-72.85$} \\

\textbf{Pairwise outcome}
& 4 wins, 5 losses
& 9 wins, 0 losses \\

\textbf{Behavior}
& Multiplies two values treated as one even and one odd, without enforcing the first-in-list constraint.
& Finds the first even and first odd elements in the list and returns their product. \\

\textbf{Induced instruction}
& ``Return the product of two numbers, one even and one odd.''
& ``Return the product of the first even and first odd numbers in the input list.'' \\
\bottomrule
\end{tabular}

\caption{Qualitative example from MBPP+ task 784 with Llama-3-8B. Coder and CR select a more generic candidate that drops the explicit first-element and given-list constraints, whereas CodeRSA selects a candidate that preserves them. CR denotes CoderReviewer.}
\label{tab:qual_case_task784}
\end{table*}
\subsection{Qualitative Analysis}
\label{app:case_study}
To complement the aggregate results, we examine a representative example from MBPP+ (task 784) with Llama-3-8B: ``Write a function to find the product of first even and odd number of a given list.''
\citet{zhang2022coderreviewerrerankingcode} note that reranking based on cumulative token likelihood tends to prefer more generic candidates, and this tendency is clearly visible here.
Both Coder and CoderReviewer select a very simple candidate whose behavior is essentially just a generic multiplication function: it takes two inputs named even and odd and returns their product.
Although this candidate is superficially related to the task, it drops two explicit constraints in the original instruction, namely that the numbers should be the first even and odd elements and that they should be taken from the given list.

This example is useful because the candidate selected by Coder and CoderReviewer is not unrelated to the task, but too generic.
Once such a candidate enters pairwise comparison, it can receive relatively strong support not only from the original instruction \(I_0\), but also from many induced instructions, because its behavior remains compatible with a broader range of interpretations.
As a result, it is less effective at distinguishing the user's intended meaning from nearby alternatives.
This is reflected in its pairwise outcome: the candidate selected by Coder and CoderReviewer wins only 4 of the 9 comparisons.
This also helps explain why the Reviewer term is insufficient in this case.
The Coder score already assigns a strong preference to the generic candidate, and the additional reverse-direction signal does not fully correct this bias, so CoderReviewer still selects the same candidate.

By contrast, the candidate selected by CodeRSA preserves the task-defining constraints that the result should be computed from the first even and first odd elements in the given list.
Because this behavior is more specifically aligned with the user instruction, it is less broadly supported by alternative induced interpretations and therefore more discriminative in pairwise comparison.
Accordingly, it achieves a much stronger pairwise profile and wins all 9 comparisons.
It also receives the higher final CodeRSA score, even after the additional global-support correction from \textit{avg}.
\begin{table*}[!t]
\centering
\small
\begin{tabular*}{\textwidth}{@{\extracolsep{\fill}}llccccc@{}}
\toprule
Dataset & Model & CR & CWUCS & Prompt A (default) & Prompt B & Prompt C \\
\midrule
\multirow{2}{*}{HumanEval+}
& Llama-3-8B & 44.5 & 42.9 & 47.8 & 45.8 & 47.1 \\
& Qwen2.5-7B & 52.0 & 55.4 & 59.9 & 60.5 & 57.9 \\
\midrule
\multirow{2}{*}{MBPP+}
& Llama-3-8B & 40.5 & 41.2 & 44.6 & 43.1 & 43.6 \\
& Qwen2.5-7B & 73.8 & 72.4 & 76.5 & 74.4 & 74.6 \\
\bottomrule
\end{tabular*}
\caption{Sanity check on induced-instruction prompt variation. Numbers report mean accuracy over five candidate-generation seeds (42--46). Prompt A is the default HumanEval+/MBPP+ prompt used in the main paper; Prompts B and C are alternative phrasings with the same behavior-description objective. CoderReviewer (CR) and Consensus-WUCS (CWUCS) are included for reference.}
\label{tab:prompt_sanity}
\end{table*}

More generally, this example illustrates the intended mechanism of CodeRSA.
Pairwise comparison favors candidates that are not merely plausible under \(I_0\), but more uniquely supported by \(I_0\) than by the induced instructions associated with competing candidates.
In this case, the generic candidate selected by Coder and CoderReviewer is too broadly compatible to be strongly discriminative, whereas the candidate selected by CodeRSA is more specifically faithful to the full task requirement.

Consensus-WUCS also selects an incorrect candidate on this example, but we do not discuss its failure in the same detail.
Unlike Coder, CoderReviewer, and CodeRSA, its selection is driven primarily by inter-candidate agreement within the sampled pool rather than by direct competition grounded in the original instruction \(I_0\).
As a result, its failure here is less naturally interpreted in prompt-level terms, and more in terms of which candidate is most central to the internal candidate consensus.
This difference highlights the distinct inductive biases of the methods: CodeRSA remains grounded in the user instruction, whereas Consensus-WUCS is shaped more strongly by candidate-internal agreement.

\subsection{Prompt Templates for Candidate-Induced Instruction Generation}
\label{app:induced-prompt}

For each sampled candidate program, we generate one candidate-induced instruction using the same base model with greedy search.
Across benchmarks, the prompts share the same behavior-description objective: the model is asked to describe what the candidate code actually does, rather than what it might have been intended to solve.
HumanEval+ and MBPP+ use the same concise one-sentence prompt template and the same few-shot examples.
BigCodeBench uses a closely related variant with the same objective and examples of similar style, but allows multi-sentence outputs in order to accommodate more complex function behavior.
The full prompt templates are listed at the end of the appendix in Appendix~\ref{app:full-induced-prompts}.

\paragraph{Prompt variation sanity check.}
Because CodeRSA relies on candidate-induced instructions, one natural concern is whether its performance depends heavily on a specific prompt wording.
To assess this, we conduct a small-scale sanity check on HumanEval+ and MBPP+, using Llama-3-8B and Qwen2.5-7B.
Across all settings, we keep the candidate pool, scoring procedure, and reranking pipeline fixed, and vary only the prompt used for induced-instruction generation.

\paragraph{Prompt A.}
The default HumanEval+/MBPP+ prompt used in the main experiments is reproduced in Appendix~\ref{app:full-induced-prompts}.

Prompts B and C share the same few-shot examples as Prompt A and differ only in their rule block.
Relative to Prompt A, Prompt B is stricter: it requires exactly one sentence and places stronger emphasis on minimal semantic content.
Relative to Prompt A, Prompt C is slightly broader: it allows one or two short sentences and permits a second sentence only when needed to state an important behavioral constraint.

For each dataset--model pair, results are averaged over the same five candidate-generation seeds (42--46) used in the main experiments.
To contextualize the results, Table~\ref{tab:prompt_sanity} also reports the corresponding accuracies of CoderReviewer (CR) and Consensus-WUCS (CWUCS), which do not depend on the induced-instruction prompt in the same way.
The purpose of this analysis is not to optimize prompt wording exhaustively, but to verify that the gains of CodeRSA are not driven by a narrowly tuned prompt.

As shown in Table~\ref{tab:prompt_sanity}, the overall pattern is stable across prompt variants.
Across all four dataset--model settings, every prompt variant remains above both CR and CWUCS\@.
At the same time, the differences among Prompt A, B, and C are modest: the largest spread is 2.6 points on HumanEval+ with Qwen2.5-7B, while the remaining settings vary by at most 2.1 points.
The default Prompt A is the strongest variant in three of the four settings, while Prompt B is slightly stronger on HumanEval+ with Qwen2.5-7B.
No variant changes the qualitative conclusion that CodeRSA remains clearly competitive with, and typically stronger than, the main reranking baselines.
This suggests that the gains of CodeRSA are not tied to one brittle prompt formulation, but are reasonably robust under modest changes in the wording and output constraints of the induced-instruction prompt.

\subsection{Manual Analysis of Induced Instructions}
\label{app:analysis-instruction}

To better understand the candidate-induced instructions used by CodeRSA, we conduct a small-scale manual analysis.
The goal is not to evaluate instruction generation as a standalone task, nor to ask whether the model can perfectly reconstruct the original human-written specification.
Instead, the analysis is motivated by a core requirement of RSA-style reasoning: for contrastive comparison to be useful, the alternative descriptions must be sufficiently \emph{specific} to distinguish one candidate from nearby alternatives.
In other words, CodeRSA does not require induced instructions to be perfect restatements of the source task; it requires them to provide informative contrasts.

We randomly sample 80 induced instructions from each of the five candidate-generation seeds used in the main experiments, yielding 400 examples in total.
Sampling is performed jointly over all dataset--model combinations so that the sample reflects the full range of benchmarks and base models considered in the paper.

For each example, we inspect the candidate code together with its induced instruction and annotate the instruction along two dimensions.

\paragraph{Behavioral faithfulness.}
This dimension measures whether the induced instruction matches the observable behavior of the code.
We use three labels:
\emph{Correct}, if the instruction accurately describes the code's behavior;
\emph{Partially correct}, if it captures the main behavior but omits or blurs an important detail;
and \emph{Incorrect}, if it substantially misdescribes the code.

\paragraph{Specificity.}
This dimension measures whether the induced instruction is specific enough to distinguish the candidate from nearby alternatives, rather than remaining overly generic.
We again use three labels:
\emph{Specific}, if the description captures behavior that would help separate the code from plausible alternatives;
\emph{Somewhat generic}, if it is broadly correct but under-informative for contrastive use;
and \emph{Too generic}, if it is so underspecified that it would remain compatible with many nearby alternatives.

Table~\ref{tab:instruction_quality} summarizes the results.
Overall, the induced instructions appear adequate for the role they play in CodeRSA.
Most are behaviorally faithful, and only a small minority are judged outright incorrect.
Most are also annotated as specific rather than generic, suggesting that the reverse-generation step usually produces descriptions that are informative enough to support local contrast among candidates.

At the same time, the analysis shows that the induced instructions are not uniformly perfect.
The most common limitation is underspecification rather than semantic failure: many instructions capture the broad behavior of the code while omitting a detail that would make the description maximally precise.
From the perspective of RSA, however, this is not necessarily problematic.
The contrastive mechanism does not require an induced instruction to reproduce the full original task description exactly.
It only requires the induced instruction to be specific enough to function as a meaningful alternative in comparison with the original instruction and other nearby descriptions.
Viewed this way, the observed level of faithfulness and specificity is consistent with the theoretical role that induced instructions play in CodeRSA.

\begin{table}[t]
\centering
\normalsize
\begin{tabular*}{\columnwidth}{@{\extracolsep{\fill}}lrr@{}}
\toprule
Category & Count & Percent \\
\midrule
\multicolumn{3}{@{}l}{\textbf{Behavioral faithfulness}} \\
Correct & 281 & 70.3 \\
Partially correct & 96 & 24.0 \\
Incorrect & 23 & 5.8 \\
\midrule
\multicolumn{3}{@{}l}{\textbf{Specificity}} \\
Specific & 339 & 84.8 \\
Somewhat generic & 43 & 10.8 \\
Too generic & 18 & 4.5 \\
\bottomrule
\end{tabular*}
\caption{Manual analysis of induced instructions on a random sample of 400 examples drawn from the five candidate-generation seeds used in the main experiments.}
\label{tab:instruction_quality}
\end{table}

\begin{table*}[t]
\centering
\footnotesize
\setlength{\tabcolsep}{4pt}
\begin{tabular*}{\textwidth}{@{\extracolsep{\fill}}llccc@{}}
\toprule
Dataset & Model & Finite RSA & CoderReviewer & CodeRSA \\
\midrule
HumanEval+ & Llama-3-8B & 42.6 & 44.5 & 47.8 \\
HumanEval+ & Qwen2.5-7B & 50.4 & 52.0 & 59.9 \\
MBPP+ & Llama-3-8B & 41.6 & 40.5 & 44.6 \\
MBPP+ & Qwen2.5-7B & 74.4 & 73.8 & 76.5 \\
\bottomrule
\end{tabular*}
\caption{Comparison between a finite vanilla RSA baseline, CoderReviewer, and CodeRSA on four representative settings. The finite RSA baseline performs direct pragmatic normalization over the induced instruction set, CoderReviewer uses bidirectional scoring, and CodeRSA uses local pairwise contests and tournament aggregation.}
\label{tab:vanilla_rsa_compare}
\end{table*}

\subsection{Comparison with a Finite Vanilla RSA Baseline}
\label{app:vanilla-rsa}

To complement the theoretical discussion in Appendix~\ref{app:theory-rsa-codersa}, we also compare CodeRSA with a finite vanilla RSA baseline on four small-model settings: HumanEval+ and MBPP+, each with Llama-3-8B and Qwen2.5-7B.
The goal of this comparison is not to claim that the finite baseline is a faithful realization of full open-ended RSA, but to test whether direct global pragmatic normalization over the induced instruction set is empirically competitive in the sampled reranking regime considered here.

Given the sampled candidate set
\[
\mathcal{C}=\{c_1,\dots,c_n\}
\]
and induced instruction set
\[
\mathcal{I}=\{I_0,I_1,\dots,I_n\},
\]
we define a finite literal listener by normalizing candidate likelihoods within the sampled pool:
\[
P_{L_0}(c_i\mid I_j)
=
\frac{P(c_i\mid I_j)}{\sum_{k=1}^{n}P(c_k\mid I_j)}.
\]
We then define a cost-free pragmatic speaker over the induced instruction set:
\[
P_{S_1}(I_j\mid c_i)
=
\frac{P_{L_0}(c_i\mid I_j)}
{\sum_{m=0}^{n}P_{L_0}(c_i\mid I_m)}.
\]
For the pragmatic listener, we use the normalized Coder score on the sampled pool as the prior:
\[
P_{\mathrm{Coder}}(c_i\mid I_0)
=
\frac{P(c_i\mid I_0)}{\sum_{k=1}^{n}P(c_k\mid I_0)}.
\]
The final pragmatic listener is then
\[
\begin{aligned}
P_{L_1}(c_i\mid I_0)
&\propto
P_{S_1}(I_0\mid c_i) \\
&\qquad
P_{\mathrm{Coder}}(c_i\mid I_0),
\end{aligned}
\]
and the final prediction is
\[
\hat{c} = \arg\max_{c_i} P_{L_1}(c_i\mid I_0).
\]
Table~\ref{tab:vanilla_rsa_compare} reports the resulting accuracy alongside CoderReviewer and CodeRSA.
Across these four settings, the finite vanilla RSA baseline is consistently weaker than CodeRSA.
The comparison with CoderReviewer is mixed: finite RSA trails CoderReviewer on HumanEval+ (42.6 vs.\ 44.5 with Llama-3-8B, and 50.4 vs.\ 52.0 with Qwen2.5-7B), but slightly exceeds it on MBPP+ (41.6 vs.\ 40.5 and 74.4 vs.\ 73.8).
In all cases, however, CodeRSA remains stronger, with gains of 2.1--9.5 points over the finite baseline.
\begin{figure*}[!t]
\centering
\includegraphics[width=0.68\textwidth]{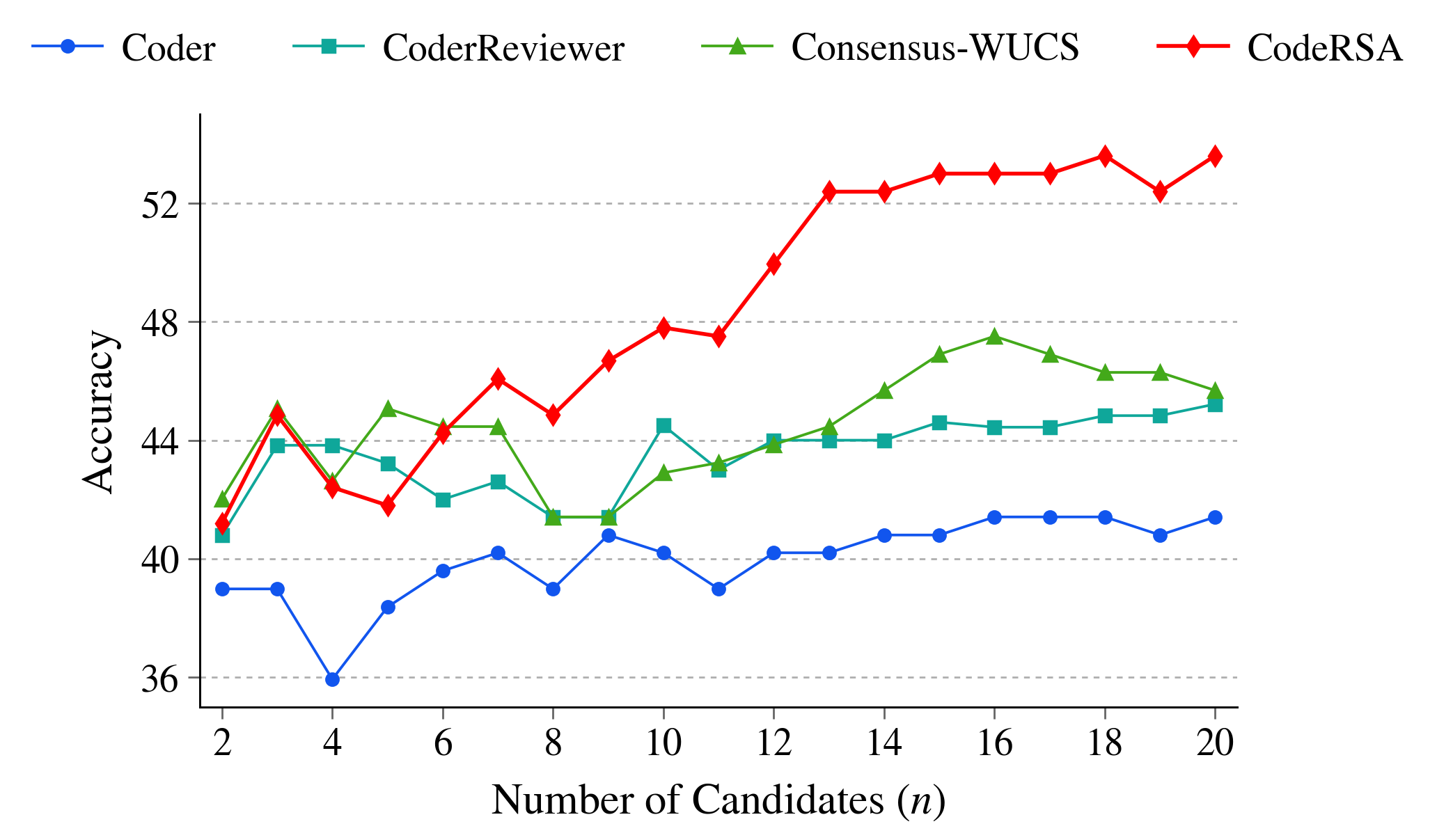}
\caption{Effect of varying the candidate-pool size \(n\) on HumanEval+ with Llama-3-8B, using a single candidate-generation seed (\(42\)). Larger candidate pools provide a richer comparison space for CodeRSA, leading to stronger performance but higher reranking cost.}
\label{fig:effect-n}
\end{figure*}

This pattern is informative in two ways.
First, it suggests that direct global pragmatic normalization over the induced instruction set can be competitive with a strong bidirectional baseline, but is less reliable than the local pairwise approximation used by CodeRSA in the sampled reranking regime considered in this paper.
Second, the comparison indicates that CodeRSA does not improve over CoderReviewer merely by introducing more RSA-like notation.
Rather, its advantage appears to come from replacing global normalization with local pairwise contests and tournament aggregation, which are better behaved in the candidate-induced comparison space.

At the same time, this comparison should be interpreted cautiously.
The finite baseline is only a tractable proxy for vanilla RSA in the sampled setting, since the true instruction space is open-ended and the induced instruction set provides only a local approximation to that space.
Accordingly, the comparison is intended mainly as an empirical illustration of the difference between direct global pragmatic normalization, bidirectional scoring, and the local pairwise approximation used by CodeRSA.

\FloatBarrier
\subsection{Effect of Candidate-Pool Size}
\label{app:candidate-number}

Because CodeRSA constructs candidate-induced instructions and performs local comparisons within the sampled candidate pool, the pool size \(n\) directly controls the richness of its comparison space.
Larger pools can provide more diverse alternatives and therefore more informative pragmatic contrasts, but they also increase the computational cost of reranking.
To examine this trade-off, we vary \(n\) on HumanEval+ with Llama-3-8B.
This analysis is a single-seed sanity check using candidate-generation seed \(42\), rather than a five-seed average as in the main results.

\begin{table*}[!t]
\centering
\footnotesize
\setlength{\tabcolsep}{3pt}
\begin{tabular*}{\textwidth}{@{\extracolsep{\fill}}llcccc@{}}
\toprule
Dataset & Model & Top-3 + CodeRSA & Top-5 + CodeRSA & Top-7 + CodeRSA & Full CodeRSA \\
\midrule
HumanEval+ & Llama-3-8B & 47.3 & 47.3 & 47.1 & 47.8 \\
HumanEval+ & Qwen2.5-7B & 56.7 & 59.8 & 60.6 & 59.9 \\
MBPP+ & Llama-3-8B & 43.9 & 45.4 & 44.7 & 44.6 \\
MBPP+ & Qwen2.5-7B & 76.8 & 77.2 & 76.8 & 76.5 \\
\bottomrule
\end{tabular*}
\caption{Mean accuracy of CodeRSA after CoderReviewer-based prefiltering to the top 3, top 5, or top 7 candidates, compared with the full 10-candidate CodeRSA pipeline. Results are averaged over five candidate-generation seeds (42--46).}
\label{tab:topk-filtering}
\end{table*}
Figure~\ref{fig:effect-n} shows that CodeRSA benefits substantially from larger candidate pools.
Although this analysis is limited to one dataset--model setting, the trend is qualitatively consistent with CodeRSA's reliance on candidate-induced local comparisons.
While the baseline rerankers improve only modestly as \(n\) increases, CodeRSA shows a clearer upward trend, rising from 42.1 at \(n=2\) to 53.4 at \(n=20\).
This pattern is consistent with the design of CodeRSA: additional candidates induce additional local instructions, which expand the comparison neighborhood and provide more opportunities for contrastive reranking.
At the same time, larger pools increase the number of candidate--instruction scores required by the method, reinforcing the cost--accuracy trade-off discussed in Section~\ref{section:Limitations}.
We therefore use \(n=10\) in the main experiments as a moderate default, while Appendix~\ref{app:top5-filtering} explores filtering as a way to reduce the effective comparison space.

\subsection{Filtering the Candidate Pool Before CodeRSA}
\label{app:top5-filtering}

To reduce the computational cost of CodeRSA, we also consider a simple two-stage variant in which CoderReviewer first preselects a smaller subset of candidates, after which CodeRSA is applied only to that subset.

We evaluate three filtered variants, keeping only the top 3, top 5, or top 7 candidates ranked by CoderReviewer before applying CodeRSA.
These are compared against the full 10-candidate CodeRSA pipeline on four small-model settings: HumanEval+ and MBPP+, each with Llama-3-8B and Qwen2.5-7B.
As elsewhere in the appendix, results report mean accuracy over the same five candidate-generation seeds (42--46) used in the main experiments.

Table~\ref{tab:topk-filtering} shows that lightweight prefiltering can preserve most of the full CodeRSA performance.
Top-3 filtering is already close to the full pipeline in three of the four settings, although it is more harmful on HumanEval+ with Qwen2.5-7B.
Top-5 and top-7 filtering are especially competitive, and in several cases slightly exceed the full 10-candidate setting, suggesting that CoderReviewer prefiltering can sometimes remove distracting candidates from the local comparison space.
Overall, these results suggest that much of the benefit of CodeRSA can be retained after a lightweight first-stage filter, provided that the retained subset remains large enough to preserve useful behavioral diversity.

\clearpage
\onecolumn
\subsection{Full Prompt Templates for Candidate-Induced Instruction Generation}
\label{app:full-induced-prompts}

This appendix lists the full prompt templates used to generate candidate-induced instructions.

\paragraph{HumanEval+ and MBPP+.}
\begin{quote}
\scriptsize
\ttfamily
Read each Python function and describe exactly what behavior it implements.\\[2pt]

Rules:\\
- Output exactly one sentence.\\
- Describe the function's actual input-output behavior only.\\
- Do not guess the intended task beyond what the code really does.\\
- Mention returned type, deduplication, ordering, and special return values when relevant.\\
- Do not write code.\\
- Do not write explanations.\\
- Do not use markdown.\\[2pt]

Function:\\
def f(xs):\\
\hspace*{1em}return list(set(xs))\\
Description:\\
Return a list of the unique elements from the input list.\\[2pt]

Function:\\
def f(xs):\\
\hspace*{1em}return sorted(set(xs))\\
Description:\\
Return a sorted list of the distinct elements in the input list.\\[2pt]

Function:\\
def f(xs, x):\\
\hspace*{1em}for i, v in enumerate(xs):\\
\hspace*{2em}if v == x:\\
\hspace*{3em}return i\\
\hspace*{1em}return -1\\
Description:\\
Return the index of the first occurrence of x in the list, or -1 if x is not present.\\[2pt]

Function:\\
\mbox{[code candidate]}\\
Description:
\end{quote}

\paragraph{BigCodeBench.}
\begin{quote}
\footnotesize
\ttfamily
Read the Python function and write a precise behavioral specification.\\[4pt]

Rules:\\
- Output two to four concise sentences.\\
- Describe only behavior that is explicitly supported by the code.\\
- Preserve observable details that tests may check: exact return type or shape, ordering, filtering, case sensitivity, labels/titles, file paths, network/file side effects, randomness/seed behavior, and exact exception or error-message behavior.\\
- Mention implementation details only when they change observable behavior.\\
- Do not guess intended behavior beyond what the code actually does.\\
- Do not return code.\\
- Do not use markdown.\\[4pt]

Function:\\
def f(groups):\\
\hspace*{1em}result = \{\}\\
\hspace*{1em}for key, values in groups.items():\\
\hspace*{2em}values = [x for x in values if x is not None]\\
\hspace*{2em}if not values:\\
\hspace*{3em}result[key] = None\\
\hspace*{2em}elif len(values) == 1:\\
\hspace*{3em}result[key] = values[0]\\
\hspace*{2em}else:\\
\hspace*{3em}result[key] = sum(values) / len(values)\\
\hspace*{1em}return result\\
Description:\\
Return a dictionary with the same keys as the input mapping. For each key, ignore None values; return None if none remain, the sole remaining value if exactly one remains, or the arithmetic mean if two or more remain.\\[4pt]

Function:\\
def f(path):\\
\hspace*{1em}with open(path, "r", encoding="utf-8") as fh:\\
\hspace*{2em}words = re.findall(r"[A-Za-z]+", fh.read().lower())\\
\hspace*{1em}counter = collections.Counter(w for w in words if len(w) >= 4)\\
\hspace*{1em}return counter.most\_common(5)\\
Description:\\
Read the UTF-8 text file and count only alphabetic words after lowercasing the entire file. Ignore words shorter than four characters, and return up to five (word, count) pairs ordered by descending frequency.\\[4pt]

Function:\\
def f(items):\\
\hspace*{1em}seen = set()\\
\hspace*{1em}out = []\\
\hspace*{1em}for x in items:\\
\hspace*{2em}if x not in seen:\\
\hspace*{3em}seen.add(x)\\
\hspace*{3em}out.append(x)\\
\hspace*{1em}return out\\
Description:\\
Return a list containing the first occurrence of each distinct input item. Preserve the original order.\\[4pt]

Function:\\
def f(url):\\
\hspace*{1em}response = urllib.request.urlopen(url)\\
\hspace*{1em}data = response.read().decode("utf-8")\\
\hspace*{1em}words = re.findall(r"\textbackslash{}b\textbackslash{}w+\textbackslash{}b", data)\\
\hspace*{1em}counts = collections.Counter(words)\\
\hspace*{1em}top = counts.most\_common(10)\\
\hspace*{1em}fig, ax = plt.subplots()\\
\hspace*{1em}ax.bar([w for w, \_ in top], [c for \_, c in top])\\
\hspace*{1em}ax.set\_xlabel("Words")\\
\hspace*{1em}ax.set\_ylabel("Frequency")\\
\hspace*{1em}ax.set\_title("Top 10 Words")\\
\hspace*{1em}return counts, ax\\
Description:\\
Fetch UTF-8 text from the URL, count regex word tokens without lowercasing or removing stopwords, and return the full Counter together with a matplotlib Axes. Plot the ten most common words as a vertical bar chart with x-axis label "Words", y-axis label "Frequency", and title "Top 10 Words".\\[4pt]

Function:\\
def f(url, download\_path):\\
\hspace*{1em}try:\\
\hspace*{2em}r = requests.get(url, stream=True)\\
\hspace*{2em}r.raise\_for\_status()\\
\hspace*{2em}if r.headers.get("Content-Type") != "application/zip":\\
\hspace*{3em}return "Error: The URL does not point to a ZIP file."\\
\hspace*{2em}with tempfile.TemporaryFile(suffix=".zip") as tmp:\\
\hspace*{3em}for chunk in r.iter\_content(chunk\_size=1024):\\
\hspace*{4em}if chunk:\\
\hspace*{5em}tmp.write(chunk)\\
\hspace*{3em}tmp.seek(0)\\
\hspace*{3em}with zipfile.ZipFile(tmp) as zf:\\
\hspace*{4em}zf.extractall(download\_path)\\
\hspace*{2em}return download\_path\\
\hspace*{1em}except zipfile.BadZipFile:\\
\hspace*{2em}return "Error: The downloaded file is not a valid ZIP file."\\
\hspace*{1em}except requests.RequestException:\\
\hspace*{2em}return "Error: Unable to download the file from the provided URL."\\
\hspace*{1em}except Exception as exc:\\
\hspace*{2em}return f"Error: \{exc\}"\\
Description:\\
Download the URL with streaming requests and require the response Content-Type to be exactly "application/zip". Extract the ZIP contents into download\_path and return download\_path on success; return the specific ZIP-type, corrupt-ZIP, download-failure, or generic error string shown by the code.\\[4pt]

Function:\\
\mbox{[code candidate]}\\
Description:
\end{quote}

\end{document}